\documentclass[journal]{IEEEtran}
\usepackage{epsfig}
\usepackage{color}
\usepackage{epstopdf}
\usepackage{graphicx}
\usepackage{amsmath}
\usepackage{amsfonts}
\usepackage{amssymb}
\usepackage{mathrsfs}
\usepackage{cite}
\usepackage{url}
\usepackage{subfigure}
\usepackage{array}
\usepackage{xcolor}
\usepackage{caption}
\usepackage{booktabs}
\usepackage{algorithm}
\usepackage{algorithmic}
\usepackage{indentfirst}
\usepackage{calc}
\usepackage{eqparbox}

\usepackage{makecell}
\usepackage{multirow}
\usepackage{setspace}

\usepackage{epsfig}
\usepackage{color}
\usepackage{epstopdf}
\usepackage{graphicx}
\usepackage{amsmath}
\usepackage{amsfonts}
\usepackage{amssymb}
\usepackage{mathrsfs}
\usepackage{cite}
\usepackage{url}
\usepackage{subfigure}
\usepackage{array}
\usepackage{xcolor}
\usepackage{caption}
\usepackage{booktabs}
\usepackage{algorithm}
\usepackage{algorithmic}
\usepackage{indentfirst}

\usepackage{makecell}
\usepackage{multirow}
\usepackage{setspace}

\usepackage[pagebackref=false,breaklinks=true,letterpaper=true,colorlinks,bookmarks=false]{hyperref}

\ifCLASSINFOpdf
\else
\fi

\begin{document}
%

\title{UncTrack: Reliable Visual Object Tracking with Uncertainty-Aware Prototype Memory Network}
%
%
%

\author{Siyuan Yao,
        Yang Guo, Yanyang Yan, Wenqi Ren and
        Xiaochun Cao,~\IEEEmembership{Senior Member,~IEEE.}

\thanks{This work was supported by National Natural Science Foundation of China (No. 62402055, No. 62302480) and Shenzhen Science and Technology Program (No. KQTD20221101093559018).}

\thanks{S. Yao and Y. Guo are with School of Computer Science (National Pilot Software Engineering School), Beijing University of Posts and Telecommunications, Beijing 100876, China. (email: yaosiyuan04@gmail.com; guoyang4409@gmail.com).}

\thanks{Y. Yan is with State Key Lab of Processors, Institute of Computing Technology, Chinese Academy of Sciences, Beijing 100190, China. (email: yanyanyang@ict.ac.cn).}

\thanks{W. Ren and X. Cao are with School of Cyber Science and Technology, Shenzhen Campus, Sun Yat-sen University, Shenzhen 518107, China. (email: rwq.renwenqi@gmail.com; caoxiaochun@mail.sysu.edu.cn).}

}


\maketitle

\begin{abstract}

Transformer-based trackers have achieved promising success and become the dominant tracking paradigm due to their accuracy and efficiency. Despite the substantial progress, most of the existing approaches tackle object tracking as a deterministic coordinate regression problem, while the target localization uncertainty has been greatly overlooked, which hampers trackers' ability to maintain reliable target state prediction in challenging scenarios. To address this issue, we propose UncTrack, a novel uncertainty-aware transformer tracker that predicts the target localization uncertainty and incorporates this uncertainty information for accurate target state inference. Specifically, UncTrack utilizes a transformer encoder to perform feature interaction between template and search images. The output features are passed into an uncertainty-aware localization decoder (ULD) to coarsely predict the corner-based localization and the corresponding localization uncertainty. Then the localization uncertainty is sent into a prototype memory network (PMN) to excavate valuable historical information to identify whether the target state prediction is reliable or not. To enhance the template representation, the samples with high confidence are fed back into the prototype memory bank for memory updating, making the tracker more robust to challenging appearance variations. Extensive experiments demonstrate that our method outperforms other state-of-the-art methods. Our code is available at \url{https://github.com/ManOfStory/UncTrack}.
\end{abstract}


\IEEEpeerreviewmaketitle

\section{Introduction}

Visual object tracking (VOT) is an essential task in computer vision society with various real-world applications, such as intelligent video surveillance, autonomous driving and robotics. Given the initial annotation in the first frame, the goal of visual object tracking is to predict the target bounding boxes across the entire video. The tracker should construct discriminative target appearance to distinguish target objects from background distracters on the fly, and flexibly adapt to challenging appearance variations of the target object, \emph{e.g.}, occlusion and deformation, etc.

Over the past decades, the matching-based approaches \cite{bertinetto2016fully,DBLP:conf/cvpr/LiYWZH18,DBLP:conf/cvpr/LiWWZXY19,yao2021learning,DBLP:conf/cvpr/TangL22}  have become the mainstream solution due to their well-balanced accuracy and efficiency. These methods handle VOT from the perspective of similarity comparison, which learns a two stream network to calculate the similarity between the template and search images for target state estimation. Recent advanced approaches employ visual transformer \cite{DBLP:conf/cvpr/0020ZWL21,DBLP:conf/cvpr/0007DBPPYG22,DBLP:conf/cvpr/CuiJ0W22,DBLP:conf/eccv/GaoZMWY22,DBLP:conf/cvpr/ChenPWLH23,DBLP:conf/cvpr/XieCLLM23,DBLP:journals/tip/LiuLWZCWL24}
to enhance the feature interaction of the template and search pairs. The transformer network splits paired template-search images into local patches and embeds them into a set of visual tokens, which are mutually aggregated using multiple cross-attention blocks. By taking scaled dot products of the query-key tokens and performing softmax-normalized interaction with the value tokens, it can capture discriminative features for accurate target object localization.

\begin{figure}[!t]
\centering

\includegraphics[ width=3.5in]{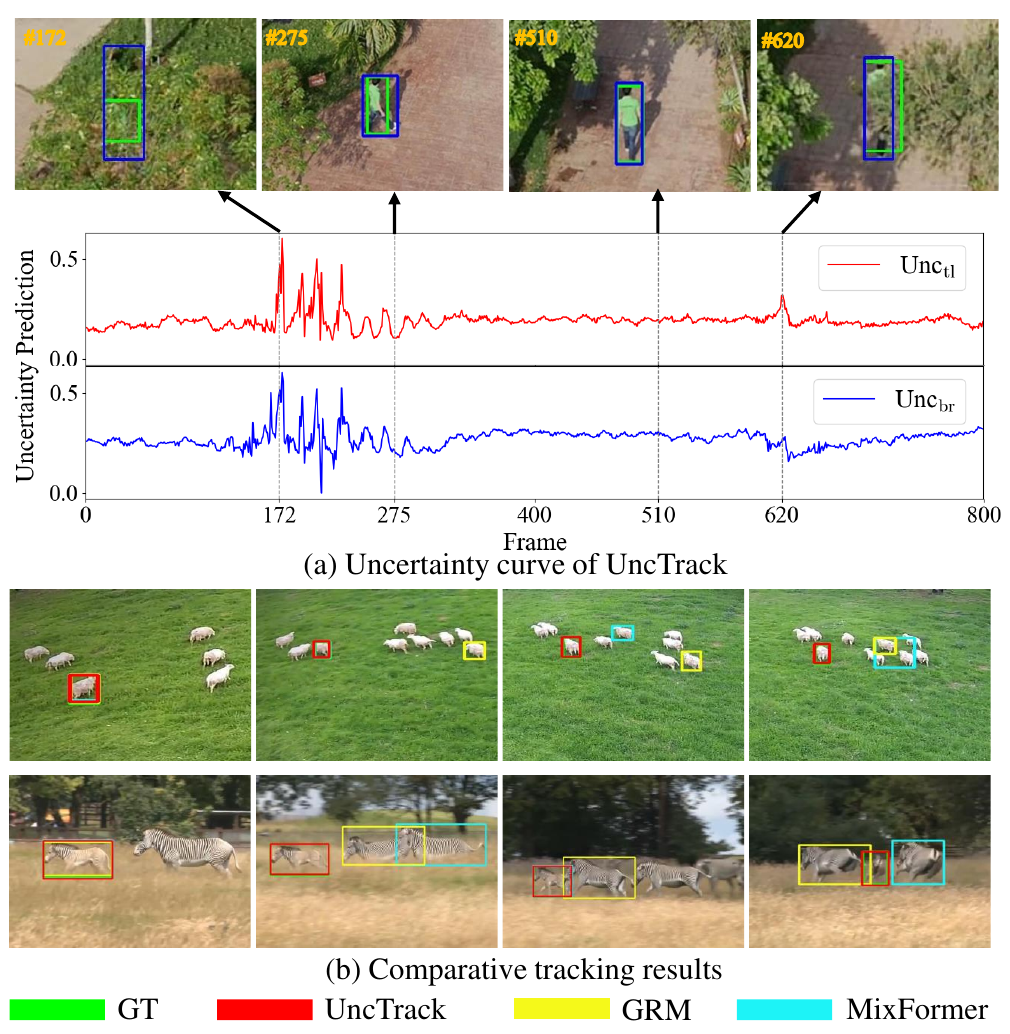}
\caption{A general introduction to our approach. (a) The uncertainty curves of the top-left and bottom-right corners predicted by the proposed UncTrack in challenging video sequence, the object occlusion raises significant localization uncertainty. The green and the blue boxes denote the ground-truth and the predicted bounding boxes, respectively. (b) The comparative tracking results of our method with other state-of-the-art trackers.}
\label{fig:motivation}
\end{figure}

Despite the promising progress, both CNN and transformer-based trackers still suffer from two drawbacks. First, existing approaches generally tackle object tracking as a deterministic coordinate regression problem, while the localization uncertainty has been overlooked or poorly investigated. As shown in Fig. \ref{fig:motivation}, the target object often encounters heavy occlusion or background distractors, which raises significant localization uncertainty in complex scenarios. Although some approaches like UAST \cite{DBLP:conf/icml/0002FZ22} and UAF \cite{DBLP:journals/tcsv/MaLZLTLJ23} employ distributional regression representation and Monte-Carlo dropout techniques to model the localization uncertainty, how to effectively leverage the uncertainty output for reliable target estimation remains unclear. Second, as the target object frequently undergoes drastic appearance variations over time, the unreliable prediction results may be accumulated across the entire video. Some prior works such as LTMU \cite{DBLP:conf/cvpr/DaiZWLLY20}, Siam R-CNN \cite{DBLP:conf/cvpr/VoigtlaenderLTL20} and KeepTrack \cite{DBLP:conf/iccv/0007DPG21} introduce online meta-updater, object re-detection or online appearance association mechanism to alleviate this problem. However, these approaches rely on local smoothness assumption to maintain temporal correspondence of the tracking object, which may easily fall into suboptimal solution that misguides the target appearance matching, yielding tracking failures eventually.

To overcome these issues, in this paper, we propose a novel uncertainty-aware transformer tracker termed UncTrack, which predicts localization uncertainty and effectively utilizes uncertainty information for accurate target state inference. Unlike previous methods \cite{DBLP:conf/iccv/0002PF0L21,DBLP:conf/cvpr/CuiJ0W22,DBLP:conf/nips/CuiSWW23} that directly predict the relative offsets or corner point coordinates of the target bounding boxes, UncTrack encodes the localization uncertainty across continuous video frames as a prototype representation, the position-aware prototypes are updated within the prototype memory bank using a temporal read-write operation, leading to more robust motion prediction in challenging scenarios. Specifically, UncTrack flattens the paired template-search regions into token patches, and performs feature interaction using a transformer encoder to establish discriminative feature representation. The output features are passed into an uncertainty-aware localization decoder (ULD), which coarsely predicts the target state and the corresponding localization uncertainty. The localization uncertainty can be regarded as an indicator to model the reliability of the predicted bounding box corners. Afterwards, the uncertainty output and the encoded template-search features are jointly fed into a prototype memory network (PMN), which exploits historical information as prototype memory bank to identify whether the target state prediction is reliable or not. Finally, the samples with high confidence are fed back into the prototype memory bank for memory updating, and we resample the template during online tracking to adapt to the target object's appearance variations. With these designs, UncTrack achieves superior performance and sets a new state-of-the-art on six prevailing benchmarks covering various complex scenarios.

In summary, the main contributions of this work can be concluded in three aspects:

$\bullet$ We design an uncertainty-aware transformer tracker called UncTrack, which is capable of exploiting localization uncertainty to maintain reliable target state prediction in challenging scenarios.

$\bullet$ We propose an uncertainty-aware localization decoder (ULD) and a prototype memory network (PMN) to predict the localization uncertainty and reliability, allowing UncTrack to identify whether the target state prediction is reliable.

$\bullet$ We update the prototype memory bank and resample the template during online tracking to adapt to the target object's appearance variations. Extensive experiments demonstrate that UncTrack outperforms existing state-of-the-art methods.

\section{Related Work}
\subsection{Transformer Based Trackers}

In the past decade, the Siamese networks have achieved significant advances in visual object tracking field. The pioneering work, \emph{i.e.} SiamFC \cite{bertinetto2016fully} uses a fully convolutional Siamese network to extract discriminative features of the template-search pairs, then employs a correlation operation to calculate the appearance similarity between the paired images. Since then, some extensive efforts also introduce region proposal branch \cite{DBLP:conf/cvpr/LiYWZH18,DBLP:conf/cvpr/LiWWZXY19}, data-augmentation strategy \cite{DBLP:conf/eccv/ZhuWLWYH18,yao2021robust} and anchor-free pipeline \cite{DBLP:conf/cvpr/GuoWC0C20,DBLP:conf/cvpr/ChenZLZJ20,DBLP:conf/eccv/ZhangPFLH20} for further improvement. However, the CNN-based Siamese models fail to effectively exploit the global context, thus they may easily fall into the local optimum.

To address this issue, the transformer architecture \cite{DBLP:conf/cvpr/ChenYZ0YL21,DBLP:conf/cvpr/0020ZWL21,DBLP:conf/cvpr/0007DBPPYG22,DBLP:conf/eccv/GaoZMWY22,DBLP:conf/cvpr/CuiJ0W22,DBLP:conf/cvpr/ChenPWLH23,DBLP:journals/tip/LiuLWZCWL24} has been introduced to aggregate the global information from sequential inputs. Chen \emph{et~al.} \cite{DBLP:conf/cvpr/ChenYZ0YL21} propose TransT, which combines the template and search region features using dual self-attention and cross-attention modules to replace the regular correlation operation. Wang \emph{et~al.} \cite{DBLP:conf/cvpr/0020ZWL21} design a hybrid CNN-Transformer architecture to encode the temporal context of target templates, such that the features across multiple video frames can be augmented by the attention mechanism. ToMP \cite{DBLP:conf/cvpr/0007DBPPYG22} employs a transformer to learn the temporal convolutional kernel to construct long-term target representation. Afterwards, to enhance the target-background discriminability in complex scenarios, researchers further propose pure transformer architecture. Cui \emph{et~al.} \cite{DBLP:conf/cvpr/CuiJ0W22} integrate target information into the feature extraction processing using a Mixed Attention Module (MAM), yielding the model to pay more attention to the target object via effective template-search feature interaction. OSTrack \cite{DBLP:conf/eccv/YeCMSC22} unifies feature extraction and relation modeling into a one-stream framework, and utilizes an early candidate elimination module to discard the irrelevant background tokens. SeqTrack \cite{DBLP:conf/cvpr/ChenPWLH23} converts the four bounding box values into a sequence of discrete tokens and uses an encoder-decoder transformer to directly predict the sequential coordinates. However, all the aforementioned transformer trackers handle object tracking as a deterministic coordinate prediction task, while the localization uncertainty has been greatly overlooked. In this work, we demonstrate that the uncertainty information is beneficial to maintain reliable target state inference in complex scenarios.



\subsection{Uncertainty Estimation}

Uncertainty estimation has attracted increasing attention for reliable deployment of deep models in safety-critical tasks. Generally, uncertainty can be divided into aleatoric uncertainty \cite{DBLP:conf/iccv/ChoiCKL19,DBLP:conf/cvpr/Wang0GFW21,DBLP:journals/tip/YaoSXWC24} and epistemic uncertainty \cite{DBLP:conf/iccv/PostelsFCNT19,harakeh2020bayesod}. The aleatoric uncertainty captures noise inherent in the data, while the epistemic uncertainty identifies the uncertainty in the model parameters. For aleatoric uncertainty, Gaussian YOLOv3 \cite{DBLP:conf/iccv/ChoiCKL19} introduces a Gaussian mixture model to predict the detection uncertainty, aiming to alleviate the target object's localization ambiguity in challenging scenarios. He \emph{et~al.} \cite{DBLP:conf/iros/HeW20} employ KL divergence loss with a minimized Gaussian distribution to predict anchor box perturbations under noisy inputs, thus the target location can be refined under the guidance of uncertainty prediction outputs. Wang \emph{et~al.} \cite{DBLP:conf/cvpr/Wang0GFW21} propose a domain adaptive segmentation network that utilizes uncertainty-based pseudo-labels for transfer learning. For epistemic uncertainty, recent advances focus on utilizing predictive probability estimation such as Bayes theories \cite{harakeh2020bayesod}, Monte-Carlo dropout \cite{MonteCarloDropout}, ensemble method \cite{DBLP:conf/nips/RahamanT21} and variational inference \cite{DBLP:conf/iccv/NakamuraOT23}. The epistemic uncertainties quantify the intrinsic trustworthiness of deep models related to various vision tasks, such as object detection \cite{DBLP:conf/nips/KendallG17}, semantic segmentation \cite{DBLP:conf/iclr/HarakehW21} and action recognition \cite{DBLP:conf/cvpr/GuoWJ22}.


Inspired by this, the uncertainty mechanism has been introduced into VOT to mitigate over-confidence. PrDiMP \cite{danelljan2020probabilistic} learns a conditional density to model the annotation noises in object tracking datasets, allowing the tracker to predict the localization ambiguities for high-quality target state refinement. UAST \cite{DBLP:conf/icml/0002FZ22} proposes a distributional regression paradigm to estimate the uncertainty, which explicitly represents the discretized probability distribution of four offsets of the target boxes. Zhou \emph{et~al.} \cite{DBLP:conf/aaai/ZhouLHLZK21} use uncertainty to select the most representative frame samples as training set, these samples are preserved during online training to enhance the discriminative power of the classifier branch. Zhuang \emph{et~al.} \cite{zhuang2022bounding} propose a CPC module, which effectively utilizes the predicted uncertainty in the bounding box distribution to calibrate the classification score and ensures accurate center point estimation. SPARK \cite{DBLP:conf/eccv/GuoXJMLXFL20} performs spatial-temporal sparse incremental perturbations on the video frames to learn the model uncertainty, aiming to degrade the tracking performance and make the adversarial attack less perceptible. Despite the progress, the aforementioned trackers pay little attention to leveraging the uncertainty prediction to maintain reliable target state estimation across multiple video frames, hence they may fail to handle the significant scene changes and object deformation. In this work, we reveal that an uncertainty-aware identification mechanism can effectively improve tracking performance.

\subsection{Memory Network}

Memory network is a type of neural network architecture that stores and retrieves information from external memory. Due to the storage flexibility and computation efficiency, the memory network has been introduced into VOT task to model the temporal correspondence. MemTrack \cite{DBLP:conf/eccv/YangC18} uses a long short-term memory network (LSTM) to read residual template information with an attention module and combines it with the initial template to update target representation. Lee \emph{et~al.} \cite{DBLP:conf/eccv/LeeCK18} further decompose the target appearance into short-term and long-term memory, then utilize a coarse-to-fine strategy to balance the states obtained by the dual memory modules. To further exploit the long-range temporal dependencies, the query-key-value (Q-K-V) structure is utilized to encode the visual semantics of both the current frame and historical frames. In \cite{fu2021stmtrack}, Fu \emph{et~al.} perform non-local matching operations between the encoded query and key features to retrieve relevant features from the space-time memory for target object localization. In \cite{DBLP:conf/cvpr/LaiLX20}, Lai \emph{et~al.} propose a two-step attention-guided memory mechanism, which helps the model learn the appearance consistency of the target object, particularly in scenarios where the object undergoes significant pose changes. Dai \emph{et~al.} \cite{DBLP:conf/cvpr/DaiZWLLY20} introduce an offline-trained meta-updater to determine whether the tracker should be updated at each timestep. Different from these classical memory-based trackers, in this work, we propose a novel prototype memory network to maintain reliable target state prediction.

\begin{figure*}
\centering
\includegraphics[width=7in]{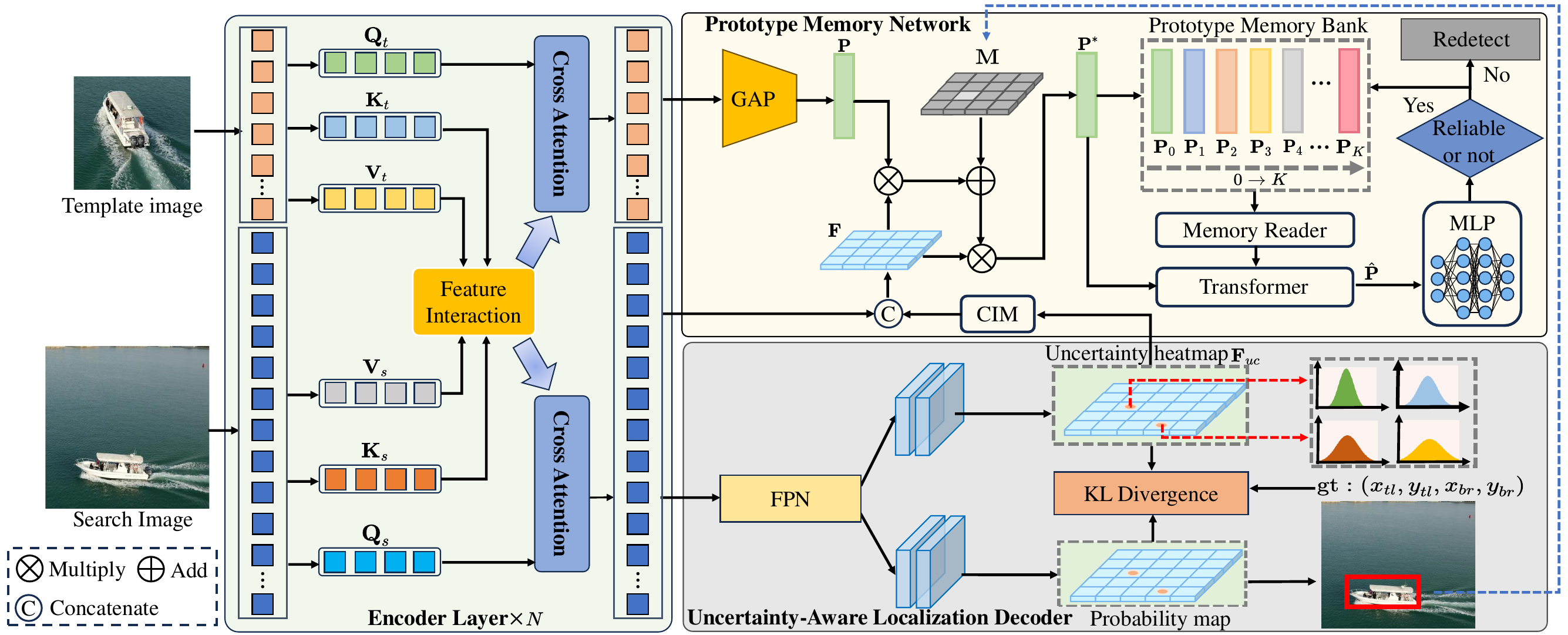}
\caption{The overall architecture of UncTrack, which consists of a transformer encoder, an uncertainty-aware localization decoder (ULD) and a prototype memory network (PMN). The paired template-search images are sent into the transformer encoder to capture the discriminative features. The encoded template-search features are further passed into ULD. The output uncertainty heatmap is transformed by the confidence inversion module (CIM) and combined with the online updated tokens in PMN, allowing UncTrack to construct reliable target-specific representation as a prototype memory bank for target state estimation.}
\label{fig:framework}
\end{figure*}

\section{Our Approach}

In this section, we present the overall architecture of the proposed UncTrack in Fig. \ref{fig:framework}. The UncTrack consists of three components: a transformer encoder, an uncertainty-aware localization decoder (ULD) and a prototype memory network (PMN). UncTrack first utilizes a transformer encoder to capture the discriminative features from paired template-search images. Then it employs ULD to coarsely predict the corner-based localization and corresponding localization uncertainty. Finally, the localization uncertainty and the encoded template-search features are sent into PMN to exploit reliable historical information,  aiming to construct reliable target-specific representation as a prototype memory bank for target state estimation.

\subsection{Transformer Backbone}

We employ a transformer encoder as feature extraction backbone with a standard Siamese pipeline. Specifically, given the paired template image $\mathbf{I}_{t} \in \mathbb{R}^{3\times H_{t}\times W_{t}}$ and the search image $\mathbf{I}_{s} \in \mathbb{R}^{3\times H_{s}\times W_{s}}$, we split the template-search images into a series of patches with the size $S\times S$, thus the template image and the search image are converted to be $ {N_{t}\times3S^{2}}$ and ${N_{s}\times 3S^{2}}$, where $N_{t} = H_{t} W_{t}/S^{2}$ and $N_{s} = H_{s}W_{s}/S^{2}$ denote the number of patches, respectively. The input patches are flattened into 1-dim vector and projected to produce the query, key and value tokens. Suppose $\mathbf{Q}_{t}$, $\mathbf{K}_{t}$ and $\mathbf{V}_{t}$ represent the embedded template tokens, $\mathbf{Q}_{s}$, $\mathbf{K}_{s}$ and $\mathbf{V}_{s}$ represent the search tokens. To model the dependencies between the template and search images, we concatenate the key-value tokens of the dual inputs for feature integration, which can be formulated as:
\begin{equation}
    \mathbf{K}= \mathrm{Concat}(\mathbf{K}_{t},\mathbf{K}_{s}), \mathbf{V}=\mathrm{Concat}(\mathbf{V}_{t},\mathbf{V}_{s}),
\end{equation}
where $\mathbf{K}$, $\mathbf{V}$ denote the combined keys and values tokens. Afterwards, the correlation between the template and search images is learned via cross-attention operation:


\begin{equation}\label{eq:Attention-T}
   \mathbf{A}_{t}=\mathrm{Softmax}(\frac{\mathbf{Q}_{t}\mathbf{K}^{T}}{\sqrt{d}} )\mathbf{V},
\end{equation}

\begin{equation}\label{eq:Attention-S}
   \mathbf{A}_{s}=\mathrm{Softmax}(\frac{\mathbf{Q}_{s}\mathbf{K}^{T}}{\sqrt{d}} )\mathbf{V},
\end{equation}
where $\mathbf{A}_{t}$ and $\mathbf{A}_{s}$ denote the output attention feature maps of the paired template-search input. $d$ denotes the channel dimension of the tokens. The whole encoder network is composed of $N$ layers to generate discriminative feature representations $\mathbf{F}_{t}$ and $\mathbf{F}_{s}$ of the template and search images, respectively. By integrating template-search token features across multiple layers, the encoder network can capture discriminative visual information regardless of the drastic appearance variations.

\subsection{Localization with Corner based Uncertainty}

The proposed UncTrack uses an uncertainty-aware localization decoder (ULD) to model the localization uncertainty of bounding box corners. Let $\mathbf{B }=\left ( {x}_{tl}, {y}_{tl}, {x}_{br}, {y}_{br} \right )$ be the corner representation of a bounding box, where $\left ( {x}_{tl}, {y}_{tl} \right )$ denotes the top-left corner and $\left ( {x}_{br}, {y}_{br} \right ) $ denotes the bottom-right one. In the following discussion, the coordinate of a bounding box $\mathbf{B}$ is denoted as $x$ for simplicity since we can optimize each coordinate independently. We send the encoded search image feature $\mathbf{F}_{s}$ into a feature pyramid network (FPN) so that the feature can be upsampled to higher resolution, then the upsampled features are disentangled into a corner localization branch and an uncertainty prediction branch. For the corner localization branch, we adopt a series of Conv-BN-ReLU layers to generate a $2\times H\times W$ probability map to estimate the position of the top-left and bottom-right corners. In the uncertainty prediction branch, the ULD predicts an uncertainty heatmap $\mathbf{F}_{uc}$ of size $4\times H\times W$, where each channel indicates the localization uncertainty of the four corner coordinates. The localization uncertainty is modeled using a univariate Gaussian distribution, which can be given by:

\begin{equation}
    P_{\theta}(x)=\frac{1}{\sqrt{2\pi}\sigma } \exp(\frac{-(x-\mu)^{2}}{2\sigma^{2}} ),
\label{eq:kl_loss}
\end{equation}
where $\theta$ denotes the learnable parameters. $\mu$ denotes the predicted corner regression, $\sigma$ is the standard deviation of the predicted corners. Note that the standard deviation $\sigma$ also measures the corner's localization uncertainty. If $\sigma \to 0$, it means the model is extremely confident about the predicted corner point's location. The ground-truth bounding box can also be formulated as a simple Dirac delta distribution:


\begin{equation}
    P_{gt}(x)=\delta(x-\mu_{gt}),
\label{eq:dirac}
\end{equation}
where $\mu_{gt}$ is the ground truth localization of the box corner point. To predict the parameter $\theta$, we minimize the KL-divergence between $P_{gt}(x)$ and $P_{\theta}(x)$, which can be calculated as:

\begin{equation}\label{eq:kl_min}
   \theta= \arg\min_{\theta} \mathrm{KL}(P_{gt}(x)\parallel P_{\theta}(x)).
\end{equation}

The KL-divergence in Eq. \ref{eq:kl_min} measures the probabilistic discrepancy between the predicted localization uncertainty and ground truth. Therefore, we introduce the uncertainty loss function for bounding box regression:
\begin{equation}
\begin{aligned}
    L_{uc}&=\mathrm{KL}(P_{gt}(x)\parallel P_{\theta}(x))\\
                     &=\int P_{gt}(x)\log P_{gt}(x)dx-\int P_{gt}(x)\log P_{\theta}(x)dx\\
                     &=\frac{(\mu-\mu_{gt})^{2}}{2\sigma^{2}} +\frac{\log(\sigma^{2})}{2} +\frac{\log(2\pi)}{2} -H(P_{gt}(x)),
    \end{aligned}
\label{eq:kl_loss}
\end{equation}
where $H(P_{gt}(x))$ is the entropy of $P_{gt}(x)$. Since $H(P_{gt}(x))$ and $\frac{\log(2\pi)}{2}$ do not depend on the estimated parameters, the loss function can be simplified as:

\begin{equation}
    L_{uc}\propto \frac{(\mu-\mu_{gt})^{2}}{2\sigma^{2}} +\frac{\log(\sigma^{2})}{2}.
\label{eq:kl_loss_new}
\end{equation}

For UncTrack, the parameters $\mu$ and $\sigma$ are predicted by the corner localization branch and uncertainty prediction branch in ULD, respectively. The loss function $L_{uc}$ in Eq. \ref{eq:kl_loss_new} indicates that the goal of ULD is to minimize the localization uncertainty and regress the corner point as closely as possible to the ground truth. If the target state is estimated inaccurately, the tracker is prone to predict larger uncertainty $\sigma$, leading $L_{uc}$ to be lower.



\subsection{Prototype Memory Network}
Although UncTrack is capable of predicting the localization uncertainty of corners, how to effectively utilize the uncertainty information to maintain reliable target state estimation remains unclear. To address this issue, we propose a Prototype Memory Network (PMN), which exploits historical template representation for reliable localization.

\noindent\textbf{Prototype Representation.} To construct prototype representation of the target object, we first squeeze the template feature $\mathbf{F}_{t}\in \mathbb{R}^{C\times H_{t}\times W_{t} }$ using global average pooling:

\begin{equation}
    \mathbf{P} = \mathrm{GAP}(\mathbf{F}_{t}),
\end{equation}
where $\mathrm{GAP}(\cdot)$ denotes the global average pooling operator. The output $\mathbf{P}\in \mathbb{R}^{C\times 1 \times 1}$ can be regarded as a compact prototype representation, which encodes the channel-wise weights of the target object. Then we convert the uncertainty map $\mathbf{F}_{uc} \in \mathbb{R}^{4\times H\times W }$ to be a certainty features $\mathbf{F}_{c} \in \mathbb{R}^{1\times H\times W }$ through a confidence inversion module (CIM), which is given by:

\begin{equation}
    \mathbf{F}_{c} = \mathrm{Sigmoid}(\mathrm{Conv}(1-\mathbf{F}_{uc})),
\end{equation}
here the function $\mathrm{Conv}(\cdot)$ is the Conv-BN-ReLU layer. Conversely, the output certain feature $\mathbf{F}_{c}$ measures the spatial localization reliability in the embedded feature domain. For each pixel in $\mathbf{F}_{c}$, the higher response indicates that the corresponding position is more likely to belong to the foreground target object with high confidence. Then it is concatenated with the embedded search image features $\mathbf{F}_{s}$ and passed through a convolutional layer, which encodes the visual semantics of the search image and the localization confidence as:

\begin{equation}
    \mathbf{F}= \mathrm{Conv}_{1\times 1}(\mathrm{Concat}(\mathbf{F}_{c},\mathbf{F}_{s})),
\end{equation}
where $\mathrm{Conv}_{1\times 1}$ is a $1\times 1$ convolutional layer aiming to recover the size of the feature channel. To enhance the foreground region, we generate a target-specific mask map $\mathbf{M} \in \mathbb{R}^{1\times H\times W }$ by labeling each pixel within the bounding box with $0$ and $-\infty $ otherwise. Then we multiply the prototype $\mathbf{P}$ with $\mathbf{F}$ and combine the target-specific mask to augment the feature $\mathbf{F}$, which can be calculated by:

\begin{equation}
    \mathbf{P}^{\ast }=\mathrm{Softmax}\left ( \mathbf{P}^{\top}\otimes \mathbf{F}+\mathbf{M} \right )\otimes \mathbf{F},
\end{equation}
the output $\mathbf{P}^{\ast}\in \mathbb{R}^{C\times 1\times 1}$ can be viewed as a reweighted prototype, which highlights the features in foreground region with high confidence.

\noindent\textbf{Prototype Memory Bank.} To maintain reliable localization, we construct a prototype memory bank to model the temporal dependencies of the embedded prototypes. Specifically, we take $K$ historical frames as guidance and project them to be prototypes, such that a prototype memory bank $\mathcal{M}=\left \{ \mathbf{P}_{0},\mathbf{P}_{1},\cdots, \mathbf{P}_{K} \right \}$ can be established. Then we design a memory read operation to retrieve the most relevant prototypes in $\mathcal{M}$ compared to the template prototype $\mathbf{P}^{\ast}$. The cosine similarities between $\mathbf{P}^{\ast}$ and the historical prototypes are computed:

\begin{equation}
\mathrm{Sim}(\mathbf{P}^{\ast},\mathbf{P}_{i}) =\frac{\mathbf{P}^{\ast}\cdot \mathbf{P}_{i}^{\top}}{\left \| \mathbf{P}^{\ast} \right \| \left \| \mathbf{P}_{i} \right \| },
    \label{eq:similarity}
\end{equation}
where $\mathbf{P}_{i}$ is the $i$-th prototype in the prototype memory bank $\mathcal{M}$. Afterwards, we select the historical prototypes with the top-k similarity scores to measure the reliability of the template prototype. Generally, if the template prototype deviates significantly from the top-k candidate prototypes in the prototype memory bank, it indicates that the appearance variations of the template prototype and the most relevant historical prototype are relatively large, thus it would be challenging to maintain reliable template prototype for robust tracking. Therefore, we concatenate the selected top-k prototypes as a grouped prototype $\mathbf{P}_{g} \in \mathbb{R}^{C\times k}$, and perform cross-attention between $\mathbf{P}^{\ast}$ and $\mathbf{P}_{g}$ using a lightweight transformer to aggregate the prototype representation as follow:

\begin{equation}
    \hat{\mathbf{P}} =\mathbf{P}^{\ast}+\mathrm{Softmax}({\phi_{q}(\mathbf{P}^{\ast})\phi_{k} (\mathbf{P}_{g})^{\top}})\cdot \phi_{v} (\mathbf{P}^{\ast}),
    \label{eq:temporal_corr}
\end{equation}
where $\hat{\mathbf{P}}$ denotes the aggregated template prototype. The functions $\phi_{q}(\cdot)$, $\phi_{k}(\cdot)$ and $\phi_{v}(\cdot)$ are learnable linear functions project the prototypes to be query-key-value structure, respectively. The cross-attention operation in Eq. \ref{eq:temporal_corr} models the temporal correlation between the template prototype and the historically grouped prototypes. To identify the reliability, the aggregated template prototype $\hat{\mathbf{P}}$ is further forward-propagated into two MLP layers to generate a 2-dimensional confidence score, which can be trained using the cross-entropy loss:

\begin{equation}\label{eq:entropy_loss}
\begin{aligned}
L_{pro} = -\frac{1}{N_{\mathcal{B}}} \sum_{i=1}^{N_{\mathcal{B}}} [y_{i}\log p_{i}+(1-y_{i})\log (1-p_{i})],
\end{aligned}
\end{equation}
where $N_{\mathcal{B}}$ denotes the amounts of the template-search pairs in a mini batch. $p_{i}$ is the prototype confidence score, $y_{i}$ is the predefined label by sampling the paired and unpaired template-search images for model training.


\subsection{Training and Online Inference}
\noindent\textbf{Model Training.} We design a two-stage training procedure for UncTrack training. In the first stage, we train the transformer encoder and ULD. The template image is generated by sampling the target box around the ground truths in the videos, while the search image is cropped in the local region with a larger size and position disturbance. The paired template-search images are fed into the network to predict the target object's location and the localization uncertainty. We jointly adopt the CIoU loss \cite{DBLP:conf/aaai/ZhengWLLYR20}, $L_{1}$ loss and the uncertainty loss for model training, which can be given by:
\begin{equation}\label{stage1 loss}
L = \alpha L_{CIoU}+\beta L_{1}+\gamma L_{uc},
 \end{equation}
where $\alpha,\beta,\gamma$ is the hyperparameters to balance the weights of the loss terms.


For the second stage, we freeze the transformer encoder and ULD, focusing on training the PMN to maintain reliable target state estimation. We first collect a series of template images in multiple video frames to construct the prototype memory bank. To train PMN, we further generate a novel set of template-search images for prototype identification. If the template image and the search image are collected from the same video, we label them as positive samples, otherwise the template-search pairs collected from different videos are labeled as negative samples. The PMN is trained using cross-entropy loss described in Eq. \ref{eq:entropy_loss}.

\noindent\textbf{Online Tracking.} During the online tracking phase, we initialize the bounding box and crop the template image in the first frame. In the subsequent video frames, the paired template-search images $\mathbf{I}_{t}$ and $\mathbf{I}_{s}$ are split into patch tokens and fed into the network to produce the predicted target box $\mathbf{B}^{*}$ and position-aware uncertainty heatmap $\mathbf{F}_{uc}$. Afterwards, the uncertainty map $\mathbf{F}_{uc}$, the paired template-search features $\mathbf{F}_{t}$ and $\mathbf{F}_{s}$ are jointly propagated into PMN to measure the reliability of the target box at this timestep. The prototype memory bank is also constructed and updated online to adapt to the target object's appearance variations. If the confidence score exceeds a pre-defined threshold $T$, we store this prototype into the prototype memory bank with a first-in-first-out mechanism. Otherwise, the predicted target box is regarded as unreliable sample, we discard this unreliable sample and resample the reliable template image $\mathbf{I}_{t}$ at the nearest video frame as input. The search area is enlarged to twice the original region, and we use a Kalman filter to maintain the temporal motion consistency of the target object.


\section{Experiment}
In this section, we first give the detailed implementation of our algorithm and perform comparisons with other state-of-the-art trackers on multiple visual tracking benchmarks. Then we conduct ablation experiments to validate the effectiveness of our approach. Finally, we visualize the tracking results and analyze the performance of our trackers in the videos with challenging attributes.

\subsection{Implementation Details}
  
We run the proposed UncTrack using Pytorch on 4 NVIDIA GeForce RTX 3090 GPUs. For the transformer backbone, we adopt ViT \cite{DBLP:conf/iclr/DosovitskiyB0WZ21} with MAE pretraining for model initialization. According to the input size and network scale, we proposed two types of UncTrack, \emph{i.e.} UncTrack-B and UncTrack-L with different input sizes. For UncTrack-B, the input template image is reshaped to $128\times128$ and the search image is reshaped to $288\times288$. For UncTrack-L, the sizes of the template and search images are  $192\times192$ and $384\times384$, respectively. The training datasets include TrackingNet \cite{DBLP:conf/eccv/MullerBGAG18}, LaSOT \cite{DBLP:conf/cvpr/FanLYCDYBXLL19}, GOT-10K \cite{DBLP:journals/pami/HuangZH21} and COCO \cite{DBLP:conf/eccv/LinMBHPRDZ14}.

In the first stage, we collect paired template-search images and train the transformer encoder and uncertainty-aware localization decoder for 500 epochs. The learning rate is $4\times 10^{-4}$ and is decreased with weight decay $1 \times 10^{-4}$. The hyperparameters $\alpha, \beta, \gamma$ are set to $2$, $5$, $2$, respectively. In the second stage, we freeze the weights of parameters pretrained in the first stage and train the prototype memory network for another 50 epochs. During this stage, we set the learning rate $5\times 10^{-5}$ and the weight decay is $1 \times 10^{-4}$. Both two stages optimize the model with ADAM. The size of the prototype memory bank is set to $6$, the threshold $T$ is set to $0.5$.

\subsection{Comparison with state-of-the-art}

\noindent\textbf{LaSOT.} LaSOT \cite{DBLP:conf/cvpr/FanLYCDYBXLL19} is a large-scale dataset containing 1,120 sequences in the training set and 280 sequences in the test set, the average length of each video is about 2,500 frames. In this experiment, we adopt normalized precision and success rate as metrics for performance evaluation.

\begin{figure}[]
\centering
\includegraphics[width=3.5in]{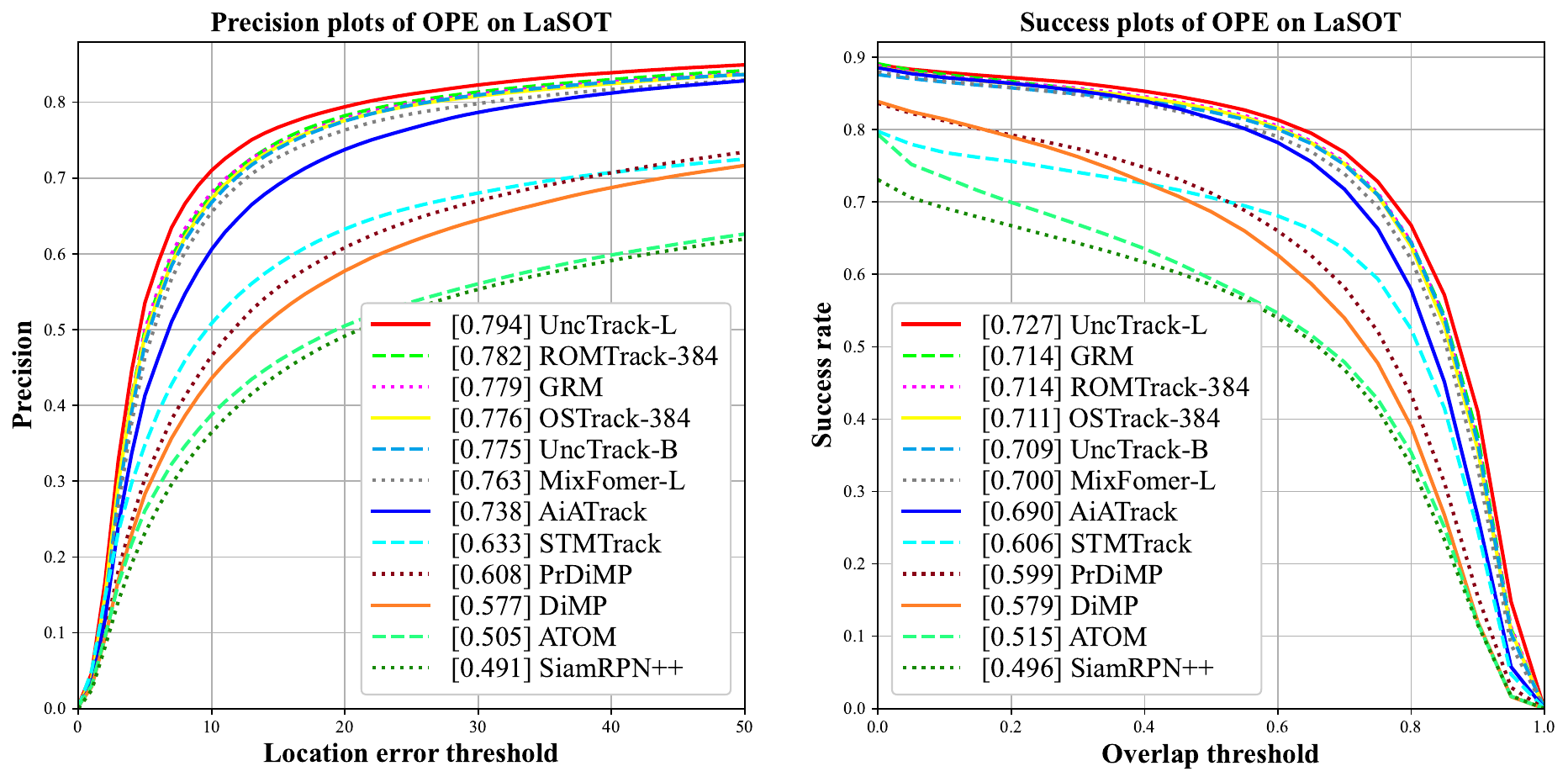}
\caption{Comparison of the proposed algorithm and several state-of-the-art trackers on LaSOT benchmark. We evaluate the precision and success rate using one-pass evaluation (OPE). }
\label{fig:LaSOT_results}
\end{figure}

Fig. \ref{fig:LaSOT_results} demonstrates the comparative results on LaSOT dataset. Our UncTrack-L obtains the precision score of $79.4\%$ and AUC score of $72.7\%$. Compared to the recent state-of-the-art tracker GRM and OSTrack-384, UncTrack-L achieves precision performance gains of $1.5\%$ and $1.8\%$. For the AUC score, UncTrack-L improves the performance of $1.3\%$ and $1.6\%$, respectively. Besides, UncTrack-B also outperforms MixFormer-L and AiATrack in both precision and success plots. Since the proposed UncTrack leverages uncertainty information to evaluate tracking reliability, it can obtain more accurate target state estimation.

\noindent\textbf{TrackingNet.} TrackingNet \cite{DBLP:conf/eccv/MullerBGAG18} is a widely used short-term tracking benchmark, which collects over 30K videos with more than 14 million dense bounding box annotations. The test set contains $511$ video sequences without publicly available ground truths, hence the results should be submitted to an online server for fair comparison. In this dataset, we adopt precision, normalized precision, and success as performance metrics.

We evaluate the proposed UncTrack on TrackingNet dataset with 14 state-of-the-art tracking methods, including AQATrack \cite{DBLP:journals/corr/abs-2403-10574}, ARTrack \cite{artrack} and SeqTrack \cite{DBLP:conf/cvpr/ChenPWLH23}, etc. The overall performance is shown in Table \ref{tab:compare-trackingnet}. Our UncTrack-L achieves the best results across all three metrics, with AUC, normalized precision, and precision scores of $86.0\%$, $90.4\%$, and $86.1\%$, respectively. UncTrack-B achieves a normalized precision score of $89.6\%$ and a success rate score of $85.0\%$. Although ARTrack-L achieves the second best score in the success rate and precision metrics due to the auto-regressive spatio-temporal prompts to propagate the motion dynamics. It lacks an effective uncertainty prediction mechanism to maintain the reliable tracking outputs.

\begin{table}
\scriptsize
\renewcommand\arraystretch{1.4}
\centering
\small{
\caption{Comparison with the state-of-the-art trackers on TrackingNet. The numbers highlighted in \textcolor[rgb]{1,0,0}{red}, \textcolor[rgb]{0,1,0}{green}, and \textcolor[rgb]{0,0,1}{blue} color stand for the result ranks at the first, second, and third place, respectively.}
\label{tab:compare-trackingnet}
\begin{tabular}{p{2.8cm}<{\centering}p{0.8cm}<{\centering}p{1cm}<{\centering}p{1cm}<{\centering}p{0.8cm}<{\centering}}
\Xhline{1pt}
Tracker & Year & AUC$(\%)$  & P$_{\textrm{Norm}}(\%)$ & P$(\%)$   \\\hline
UncTrack-L & \textbf{-} & \textcolor[rgb]{1,0,0}{86.0} & \textcolor[rgb]{1,0,0}{90.4} & \textcolor[rgb]{1,0,0}{86.1} \\
UncTrack-B & \textbf{-} & 85.0 & \textcolor[rgb]{0,0,1}{89.6} & 84.5 \\ \hline
AQATrack-384 \cite{DBLP:journals/corr/abs-2403-10574} & 2024 & 84.8 & 89.3 & 84.3 \\
SRTrack\cite{DBLP:journals/tip/LiuLWZCWL24} & 2024 & 83.8 & 88.1 & 82.9 \\
EVPTrack \cite{DBLP:conf/aaai/ShiZLLZL24} & 2024 & 84.2 & 89.1 & 84.1\\
ARTrack-L \cite{artrack} & 2023 & \textcolor[rgb]{0,1,0}{85.6} & \textcolor[rgb]{0,0,1}{89.6} & \textcolor[rgb]{0,1,0}{86.0}\\
SeqTrack \cite{DBLP:conf/cvpr/ChenPWLH23} & 2023 & \textcolor[rgb]{0,0,1}{ 85.5} & \textcolor[rgb]{0,1,0}{89.8} & \textcolor[rgb]{0,0,1}{85.8} \\
VideoTrack \cite{DBLP:conf/cvpr/XieCLLM23} & 2023 & 83.8 & 88.7 & 83.1 \\
ROMTrack-384 \cite{cai2023robust} & 2023 & 84.1 & 89.0 & 83.7 \\
GRM \cite{DBLP:conf/cvpr/GaoZZ23} & 2023 & 84.0 & 88.7 & 83.3 \\
MixFormer-L \cite{DBLP:conf/cvpr/CuiJ0W22} & 2022 & 83.9 & 88.9 & 83.1\\
OSTrack-384 \cite{DBLP:conf/eccv/YeCMSC22} & 2022 & 83.9 & 88.5 & 83.2 \\
STMTrack \cite{fu2021stmtrack} & 2021 & 80.3 & 85.1 & 76.7 \\
KYS \cite{DBLP:conf/eccv/BhatDGT20} & 2020 & 74.0 & 80.0 & 68.8 \\
DiMP \cite{DBLP:conf/iccv/BhatDGT19} & 2019 & 74.0 & 80.1 & 68.7 \\
ATOM \cite{DBLP:conf/cvpr/DanelljanBKF19} & 2019 & 70.3 & 77.1 & 64.8 \\
SiamRPN++ \cite{DBLP:conf/cvpr/LiWWZXY19}  & 2019 & 73.3 & 80.0 & 69.4 \\
\Xhline{1pt}
\end{tabular}
}
\end{table}

\begin{figure}[]
\centering
\includegraphics[width=3.5in]{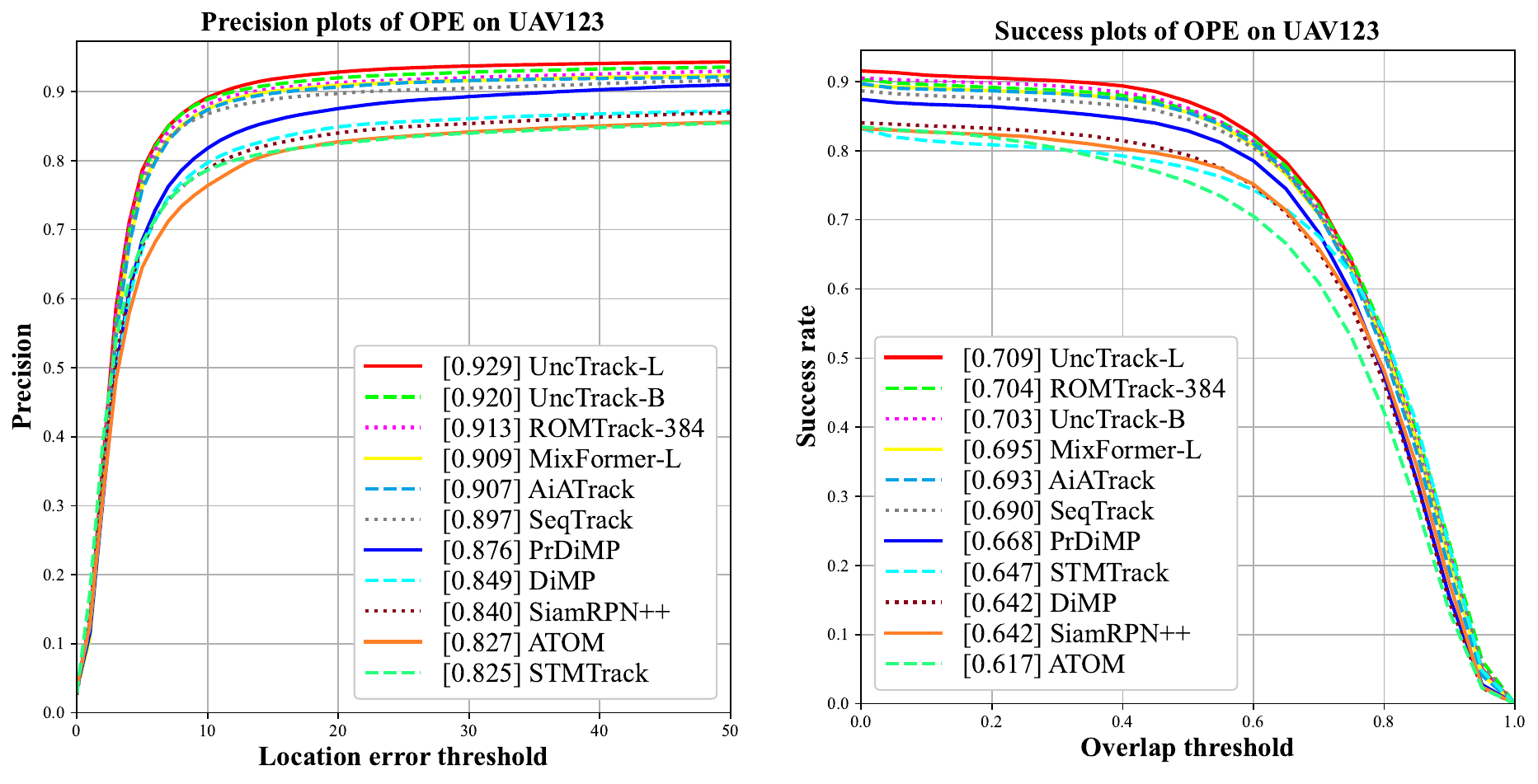}
\caption{Comparison of the proposed algorithm and several state-of-the-art trackers on UAV123 benchmark. We evaluate distance precision and overlap success plots over 123 sequences using one-pass evaluation (OPE).}
\label{fig:UAV_results}
\end{figure}

\noindent\textbf{GOT-10K.} GOT-10K \cite{DBLP:journals/pami/HuangZH21} is a large-scale dataset with 563 object classes and 87 motion patterns in the real world. The test subset of GOT-10K contains 180 video sequences with 84 different object classes. To evaluate the performance, the average overlap (AO) and success rate (SR) are employed as generic metrics in the evaluation toolkit.

Tab. \ref{tab:compare-got10k} shows the comparative results on GOT-10K dataset. UncTrack-L achieves the best AO and SR$_{0.75}$ results, it also achieves the second best score in the SR$_{0.5}$ metric. Compared to existing state-of-the-art trackers like SeqTrack and ROMTrack-384, UncTrack-L obtains obvious performance gains across all three metrics. We also observe that MixFormer-L \cite{DBLP:conf/cvpr/CuiJ0W22} achieves the best SR$_{0.5}$ scores, but it fails to obtain more accurate results in SR$_{0.75}$ metric, which indicates that MixFormer-L can not precisely predict the target object's coordinates in a long time span due to the unreliable tracking results accumulation.


\noindent\textbf{UAV123.} UAV123\cite{DBLP:conf/eccv/MuellerSG16} is a dataset specifically designed for unmanned aerial vehicle (UAV) vision tasks. It contains 123 UAV flight video sequences with more than 112K
frames covering a variety of challenging aerial scenarios such as occlusion, low resolution, out-of-view, etc.

As shown in Fig. \ref{fig:UAV_results}, UncTrack-L obtains the scores of $92.9\%$ in precision and $70.9\%$ in success rate. Compared to the state-of-the-art tracker MixFormer-L, it achieves the performance improvements with $2.0\%$ and $1.4\%$ in these two metrics. The UncTrack-B also outperforms MixFormer-L, AiATrack and SeqTrack even using a smaller input size. Such results reveal that uncertainty prediction, combined with the prototype memory mechanism, helps UncTrack maintain stable tracking, especially for the tiny UAV objects.

\begin{table}
\scriptsize
\renewcommand\arraystretch{1.4}
\centering
\small{
\caption{Comparison with the state-of-the-art trackers on GOT-10K. The numbers highlighted in \textcolor[rgb]{1,0,0}{red}, \textcolor[rgb]{0,1,0}{green}, and \textcolor[rgb]{0,0,1}{blue} color stand for the result ranks at the first, second, and third place, respectively.}
\label{tab:compare-got10k}
\begin{tabular}{p{2.8cm}<{\centering}p{0.8cm}<{\centering}p{1cm}<{\centering}p{1cm}<{\centering}p{1.1cm}<{\centering}}
\Xhline{1pt}
Tracker & Year & AO$(\%)$  &SR$_{0.5}(\%)$ & SR$_{0.75}(\%)$   \\\hline
UncTrack-L & \textbf{-} & \textcolor[rgb]{1,0,0}{76.5} & \textcolor[rgb]{0,1,0}{85.4} & \textcolor[rgb]{1,0,0}{76.5} \\
UncTrack-B & \textbf{-} & \textcolor[rgb]{0,0,1}{75.6} & 85.1 & \textcolor[rgb]{0,1,0}{75.2} \\ \hline
AQATrack-384 \cite{DBLP:journals/corr/abs-2403-10574} & 2024 & \textcolor[rgb]{0,1,0}{76.0} & \textcolor[rgb]{0,0,1}{85.2} & \textcolor[rgb]{0,0,1}{74.9} \\
SRTrack\cite{DBLP:journals/tip/LiuLWZCWL24} & 2024 & 73.4 & 83.9 & 69.9 \\
STCFormer\cite{DBLP:conf/aaai/Hu0HZCRT24} & 2024 & 74.3 & 84.2 & 72.6 \\
SeqTrack \cite{DBLP:conf/cvpr/ChenPWLH23} & 2023 & 74.8 & 81.9 & 72.2 \\
VideoTrack\cite{DBLP:conf/cvpr/XieCLLM23} & 2023 & 72.9 & 81.9 & 69.8 \\
ROMTrack-384 \cite{cai2023robust} & 2023 & 74.2 & 84.3 & 72.4 \\
GRM \cite{DBLP:conf/cvpr/GaoZZ23} & 2023 & 73.4 & 82.9 & 70.4 \\
MixFormer-L \cite{DBLP:conf/cvpr/CuiJ0W22} & 2022 & 75.6 & \textcolor[rgb]{1,0,0}{85.7} & 72.8\\
OSTrack-384 \cite{DBLP:conf/eccv/YeCMSC22} & 2022 & 73.7 & 83.2 & 70.8 \\
SimTrack-L\cite{DBLP:conf/eccv/ChenLBQSLGWO22} & 2022 & 69.8 & 78.8 & 66.0 \\
STMTrack \cite{fu2021stmtrack} & 2021 & 64.2 & 73.7 & 57.5 \\
KYS \cite{DBLP:conf/eccv/BhatDGT20} & 2020 & 63.6 & 75.1 & 51.5 \\
DiMP \cite{DBLP:conf/iccv/BhatDGT19} & 2019 & 61.1 & 71.7 & 49.2 \\
ATOM \cite{DBLP:conf/cvpr/DanelljanBKF19} & 2019 & 55.6 & 63.4 & 40.2 \\
SiamRPN++ \cite{DBLP:conf/cvpr/LiWWZXY19}  & 2019 & 51.7 & 61.6 &32.5 \\
\Xhline{1pt}
\end{tabular}
}
\end{table}

\begin{figure}[]
\centering
\includegraphics[width=3.5in]{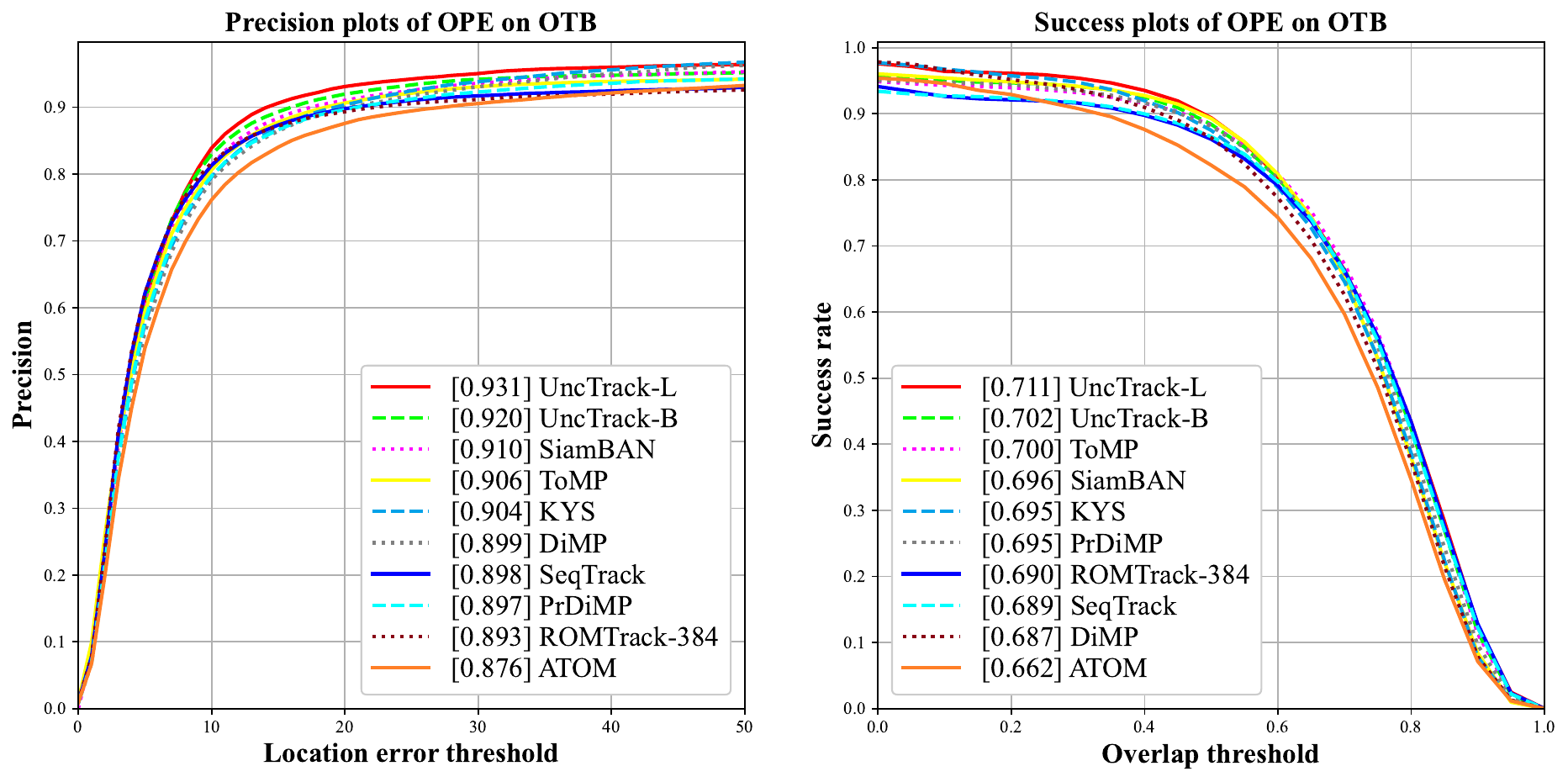}
\caption{Comparison of the proposed algorithm and several state-of-the-art trackers on OTB benchmark. We evaluate distance precision and overlap success plots over 100 sequences using one-pass evaluation (OPE).}
\label{fig:OTB_results}
\end{figure}
\noindent\textbf{VOT2020.} VOT is a challenging visual tracking benchmark that adopts a reset-based methodology to evaluate the tracking performance. The target bounding boxes of this benchmark are annotated by five attributes, including occlusion, illumination change, motion change, size change, and camera motion. The performance is measured in the terms of Accuracy, Robustness and Expected Average Overlap (EAO). The VOT2020 version \cite{DBLP:conf/eccv/KristanLMFPKDZL20} consists of 60 videos.

We report the performance of our proposed trackers on VOT2020, the experimental results are shown in Table \ref{tab:compare-vot2020}. As we can see, our UncTrack-L achieves the top-ranked performance on EAO criteria of $0.581$, and robustness of $0.891$. Our UncTrack-B achieves the top-ranked performance with the accuracy of $0.771$. Since VOT2020 utilizes re-initialization strategy to evaluate the tracking performance, our UncTrack tends to achieve better performance because the prototype memory network can alleviate the temporal drifting to avoid the undesired re-initialization.

\noindent\textbf{OTB100.} OTB100 \cite{DBLP:journals/pami/WuLY15} is a classic dataset for object tracking evaluation. It consists of 100 video sequences with various scenes and challenges, such as target deformation, occlusion, and illumination variation, etc.

We present the comparative results in Fig. \ref{fig:OTB_results}. The experiment shows that our UncTrack-L and UncTrack-B have achieved new SOTA performance on OTB100 dataset. UncTrack-L and UncTrack-B obtain the leading performance of $(0.931, 0.711)$ and $(0.920, 0.702)$ in precision and success rate plots. Both two variation trackers surpass recent state-of-the-art trackers, \emph{e.g.} ROMTrack-384 \cite{cai2023robust} and SeqTrack \cite{DBLP:conf/cvpr/ChenPWLH23}.
\begin{table}
\scriptsize
\renewcommand\arraystretch{1.2}
\centering
\small{
\caption{Comparison with the state-of-the-art trackers on VOT2020. The bold number indicates the best result in the relative evaluation metric.}
\label{tab:compare-vot2020}
\begin{tabular}{p{2.5cm}<{\centering}p{1.5cm}<{\centering}p{1.7cm}
<{\centering}p{1.5cm}<{\centering}}
\Xhline{1pt}
Tracker &EAO  &Accuracy &Robustness    \\\hline
UncTrack-L & \textbf{0.581} & 0.761 & \textbf{0.891} \\
UncTrack-B & 0.569 & \textbf{0.771} & 0.871 \\ \hline
MixFormer-L \cite{DBLP:conf/cvpr/CuiJ0W22} & 0.555 & 0.762 & 0.855 \\
OSTrack-384 \cite{DBLP:conf/eccv/YeCMSC22} & 0.524 & 0.767 & 0.816 \\
ToMP \cite{DBLP:conf/cvpr/0007DBPPYG22} & 0.497 & 0.750 & 0.798 \\
STARK \cite{DBLP:conf/iccv/0002PF0L21} & 0.505 & 0.759 & 0.817 \\
OceanPlus \cite{DBLP:journals/tip/ZhangLLHP21} & 0.491 & 0.685 & 0.842 \\
D3S \cite{DBLP:conf/cvpr/LukezicMK20} & 0.439 & 0.699 & 0.769 \\
DiMP \cite{DBLP:conf/iccv/BhatDGT19} & 0.278 & 0.457 & 0.734 \\
ATOM \cite{DBLP:conf/cvpr/DanelljanBKF19} & 0.271 & 0.462 & 0.734 \\
SiamMask \cite{DBLP:conf/cvpr/Wang0BHT19} & 0.321 & 0.624 & 0.648 \\
SiamFC \cite{bertinetto2016fully} & 0.179 & 0.418 & 0.502 \\
\Xhline{1pt}
\end{tabular}
}
\end{table}

\subsection{Ablation Analysis}
In this subsection, we conduct ablation studies on GOT-10K \cite{DBLP:journals/pami/HuangZH21} to analyse the effectiveness of the proposed UncTrack.

\begin{table}[]
\scriptsize
\renewcommand\arraystretch{1.2}
\centering
\small{
\caption{Ablation studies of the different variants in UncTrack-B on GOT-10K dataset. The bold number indicates the best result obtained in the experiment.}
\label{tab:ablation-uncpro}
\begin{tabular}{p{1.2cm}<{\centering}p{1.2cm}<{\centering}p{1.2cm}<{\centering}p{1.2cm}<{\centering}p{1.2cm}}
\Xhline{1pt}
 \multicolumn{2}{c}{Setting} & \multicolumn{3}{c}{GOT-10K}      \\
\cmidrule(r){1-2} \cmidrule(r){3-5}
 ULD & PMN  & AO & SR$_{0.5}$ & SR$_{0.75}$ \\\hline
 $\times$ &  $\times$  &  74.3   & 84.1  & 73.0  \\
 $\checkmark$ & $\times$ &  75.1  & 84.5 & 74.6 \\
 $\times$ & $\checkmark$ & 75.3 & 84.5 & 74.9 \\
 $\checkmark$ & $\checkmark$ & \textbf{75.6} & \textbf{85.1} & \textbf{75.2} \\
\Xhline{1pt}
\end{tabular}
}
\end{table}

\begin{figure}[]
\centering
\includegraphics[width=3.5in]{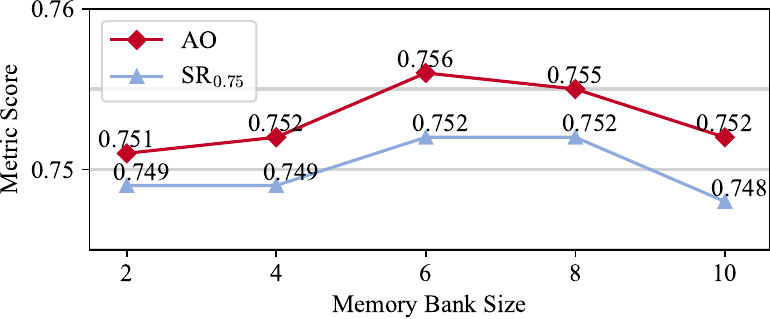}
\caption{Ablation study of the memory bank size in UncTrack with different settings.}
\label{fig:ablation-Memory_Bank_Size}
\end{figure}

\noindent\textbf{Ablation of Different Variants.}
To demonstrate the effectiveness of each module, we selectively choose the proposed Uncertainty Localization Decoder (ULD) and Prototype Memory Network (PMN) in UncTrack-B for ablation analyses. If ULD is discarded in the experiment, it will be replaced by a standard FCN localization head.

As shown in Tab. \ref{tab:ablation-uncpro}, the baseline tracker without ULD and PMN obtains the $74.3\%$ and $73.0\%$ in terms of AO and SR$_{0.75}$. By introducing ULD, the AO and SR$_{0.75}$ performance can be improved to $75.1\%$ and $74.6\%$, respectively. When PMN is incorporated into the tracking framework, the variant tracker achieves scores of $75.3\%$ and $74.9\%$ in these metrics due to its ability to memorize the historical target appearance. Furthermore, if ULD and PMN are simultaneously introduced into the framework, the variant track achieves the AO score of $75.6\%$ and the SR$_{0.75}$ score of $75.2\%$, outperforming the baseline tracker with $1.3\%$ and $2.2\%$. Such results verify the effectiveness of the proposed ULD and PMN.

\begin{table}[]
\scriptsize
\renewcommand\arraystretch{1.2}
\centering
\small{
\caption{Ablation study of the prototype memory mechanism in UncTrack-B with different settings on GOT-10K dataset. The bold number indicates the best result obtained in the experiment.}
\label{tab:ablation-memory-mechanism}

\begin{tabular}{p{2.8cm}<{\centering}p{1cm}<{\centering}p{1.3cm}<{\centering}p{1.2cm}}
\Xhline{1pt}
\multirow{2}{*}{Memory mechanism}  & \multicolumn{3}{c}{GOT-10K}      \\
\cmidrule(r){2-4}
 & AO & SR$_{0.5}$ & SR$_{0.75}$ \\\hline
STM \cite{fu2021stmtrack} &  74.9  & 84.3 & 74.1 \\
LSTM \cite{DBLP:conf/eccv/LeeCK18}  & 74.8 & 84.3 & 74.1 \\
GRU \cite{chung2014empirical} & 74.2 & 83.5 & 73.4 \\
\textbf{Ours} &  \textbf{75.6}   & \textbf{85.1}  & \textbf{75.2}  \\
\Xhline{1pt}
\end{tabular}
}
\end{table}

\noindent\textbf{Memory mechanisms.}
To verify the effectiveness of the memory mechanism, we analyse the performance changes using different memory mechanisms in UncTrack-B. We adopt STM \cite{fu2021stmtrack}, Long Short-Term Memory (LSTM) \cite{DBLP:conf/eccv/LeeCK18}, Gated Recurrent Unit (GRU) \cite{chung2014empirical} and our proposed prototype memory network for fair comparison. The comparative results are shown in Tab. \ref{tab:ablation-memory-mechanism}.

We can see that our prototype memory network obtains the best scores in all evaluation metrics. On GOT-10K benchmark, our approach achieves the performance gain of $0.7\%$, $0.8\%$, $1.1\%$ against the variant tracker using STM in terms of AO, SR$_{0.5}$ and SR$_{0.75}$, respectively. Compared to GRU, it also achieves the performance gains of $1.4\%$, $1.6\%$ and $1.8\%$ for each metric. Such results show that the prototype memory bank can provide more valuable temporal information to maintain reliable target state prediction, leading the tracker to be more robust against the appearance variations.

\noindent\textbf{Ablation Analyses on the Backbone.}
We present the ablation analyses to investigate the performance changes without MAE, the results are shown in Table \ref{tab:table_VIT}. We can see that the variant UncTrack without MAE achieves satisfying performance. For UncTrack-L, the variant tracker without MAE also obtains the AUC score $69.9\%$ and precision score $90.6\%$ on UAV123 dataset. For UncTrack-B, the performance of the variant approach without MAE decreases $1.7\%$ and $2.8\%$ in terms of AUC and precision on TrackingNet dataset, respectively.

\begin{table}[h]
\scriptsize
\renewcommand\arraystretch{1.2}
\centering
\small{
\caption{Ablation analyses of the performance changes with or without MAE. Here ``w/o MAE" means we use the pure ViT backbone.}
\label{tab:ViT_compare}
\begin{tabular}{p{3.2cm}<{\centering}p{0.6cm}<{\centering}p{0.6cm}<{\centering}p{0.6cm}
<{\centering}p{0.6cm}<{\centering}p{0.6cm}<{\centering}p{0.6cm}}
\Xhline{1pt}
 \multirow{2}{*}{Tracker} & \multicolumn{2}{c}{UAV123} & \multicolumn{3}{c}{TrackingNet}     \\
 \cmidrule(r){2-3} \cmidrule(r){4-6}
  & AUC & P & AUC  & P$_{\textrm{Norm}}$ & P\\\hline
  UncTrack-L & \textbf{70.9} & \textbf{92.9} & \textbf{86.0} & \textbf{90.4} & \textbf{86.1}\\
  UncTrack-L(w/o MAE) & 69.9 & 90.6 & 85.8 & 90.2 & 86.0 \\
  UncTrack-B & 70.3 & 92.0 & 85.0 & 89.6 & 84.5 \\
  UncTrack-B(w/o MAE) & 69.1 & 90.6 & 83.3 & 88.2 & 81.7 \\
\Xhline{1pt}
\label{tab:table_VIT}
\end{tabular}
}
\end{table}

\noindent\textbf{Size of Prototype Memory Bank.}
We conduct an ablation study to investigate whether different sizes of the memory bank will affect the proposed tracker. In this experiment, we change the size of the prototype memory bank in UncTrack-B for comparison, the results are shown in Fig. \ref{fig:ablation-Memory_Bank_Size}.

We can see that the prototype memory bank size impacts the tracking performance in both AO and SR$_{0.75}$ metrics. If the memory size is too small, UncTrack-B can not effectively establish sufficient prototype examples for appearance updating. As the prototype memory bank size becomes larger, the AO and SR$_{0.75}$ scores also improve due to the inclusion of valuable historical prototypes for reliability identification. However, when the size is too large, the noisy prototypes may be added, leading to undesired performance degradation. Therefore, we set the prototype memory bank size to be 6 to obtain the optimal results.

\begin{table}[]
    \renewcommand\arraystretch{1.2}
    \centering
    \small{
    \caption{Model Comparison of MACs and Parameters.}
    \begin{tabular}{c c c}
        \Xhline{1pt}
        Model & MACs & Params \\
        \hline
        UncTrack-B & 57.751G & 109.257M \\
        UncTrack-B (w/o ULD and PMN) & 56.362G & 95.992M \\
        UncTrack-L & 321.593G & 309.978M \\
        UncTrack-L (w/o ULD and PMN) & 317.500G & 286.790M \\
        \Xhline{1pt}
    \end{tabular}
    \label{tab:table_params_change}
    }
\end{table}

\noindent\textbf{Computation Complex Analyses.}
We report the parameters and computational complexity changes of the ULD and PMN, the quantitative analyses are shown in Table \ref{tab:table_params_change}. We observe that the model's computational complexity and parameters change slightly when ULD and PMN are included. For UncTrack-B without ULD and PMN, the MACs are 56.362G and the number of parameters is 95.992M. With the inclusion of ULD and PMN, the MACs increase slightly to 57.751G, and the parameters increase to 109.257M. For UncTrack-L, the baseline tracker without ULD and PMN requires the MACs 317.500G and the parameters are 286.790M. With ULD and PMN, the MACs increase to 321.593G, and the parameters increase to 309.978M.

\noindent\textbf{Ablation Analyses on Threshold $T$.}
We report the performance sensitivity of UncTrack to different threshold values \(T\), the experiment is conducted on VOT2020 benchmark. As shown in Table~\ref{tab:thresholdT}, the EAO and accuracy scores are not sensitive to the variations of threshold \(T\), while the robustness score would be influenced by the threshold \(T\) as it will change the updating frequency of the prototype memory bank.

\begin{table}
\scriptsize
\renewcommand\arraystretch{1.2}
\centering
\small{
\caption{Comparative performance of UncTrack-B with different threshold $T$ on VOT2020 benchmark.}
\label{tab:thresholdT}
\begin{tabular}{p{2.5cm}<{\centering}p{1.5cm}<{\centering}p{1.7cm}
<{\centering}p{1.5cm}<{\centering}}
\Xhline{1pt}
Threshold ($T$) & EAO  &Accuracy &Robustness    \\\hline

0.50 & \textbf{0.569} & 0.771 & \textbf{0.871} \\
0.70 & 0.566 & 0.773 & 0.863 \\
0.90 & 0.563 & \textbf{0.775} & 0.856 \\
\Xhline{1pt}
\end{tabular}
}
\end{table}

\noindent\textbf{Generalization Analyses.}
We present an experiment to analyse the generalization capability of the proposed UncTrack on NAT2021 benchmark. NAT2021 is a dataset collected for specific nighttime conditions. As shown in Table~\ref{tab:compare-nat2021-1}, UncTrack-B outperforms all SOTA trackers, \emph{e.g.} ODTrack-B, ARTrackV2-256, etc. Compared to ARTrackV2-256, it obtains performance gains of $2.2\%$ and $1.4\%$ in terms of precision and AUC, respectively. Compared to AQATrack-256, UncTrack-B significantly surpasses it with the precision and AUC improvement of $9.3\%$ and $6.9\%$.

\begin{table}
\scriptsize
\renewcommand\arraystretch{1}
\centering
\small{
\caption{Comparison with the SOTA trackers on NAT2021. The numbers highlighted in \textcolor[rgb]{1,0,0}{red}, \textcolor[rgb]{0,1,0}{green}, and \textcolor[rgb]{0,0,1}{blue} color stand for the result ranks at the first, second, and third place.}
\label{tab:compare-nat2021-1}
\begin{tabular}{p{2.8cm}<{\centering}p{0.8cm}<{\centering}p{1cm}<{\centering}p{1cm}<{\centering}p{0.8cm}<{\centering}}
\Xhline{1pt}
Tracker & Year & AUC$(\%)$ & P$(\%)$   \\\hline
UncTrack-B & \textbf{-} & \textcolor[rgb]{1,0,0}{61.3} & \textcolor[rgb]{1,0,0}{80.5} \\ \hline
ODTrack-B \cite{DBLP:conf/aaai/ZhengZLMZL24} & 2024 & \textcolor[rgb]{0,1,0}{60.8} & \textcolor[rgb]{0,0,1}{79.3} \\
ARTrackV2-256 \cite{DBLP:conf/cvpr/0001ZG024} & 2024 & \textcolor[rgb]{0,0,1}{59.9} & 78.3 \\
AQATrack-256 \cite{DBLP:journals/corr/abs-2403-10574} & 2024 & 54.4 & 71.2 \\
EVPTrack-224 \cite{DBLP:conf/aaai/ShiZLLZL24} & 2024 & 53.6 & 70.5 \\
MixViT-B \cite{DBLP:journals/pami/CuiJWW24} & 2024 & 58.7 & \textcolor[rgb]{0,1,0}{79.8} \\
ROMTrack-256 \cite{cai2023robust} & 2023 & 56.0 & 73.5 \\
SeqTrack-B \cite{DBLP:conf/cvpr/ChenPWLH23} & 2023 & 51.9 & 69.9 \\
OSTrack-256 \cite{DBLP:conf/eccv/YeCMSC22} & 2022 & 53.4 & 70.5 \\
\Xhline{1pt}
\end{tabular}
}
\end{table}

\begin{figure*}[]
\centering
\includegraphics[width=6.1in]{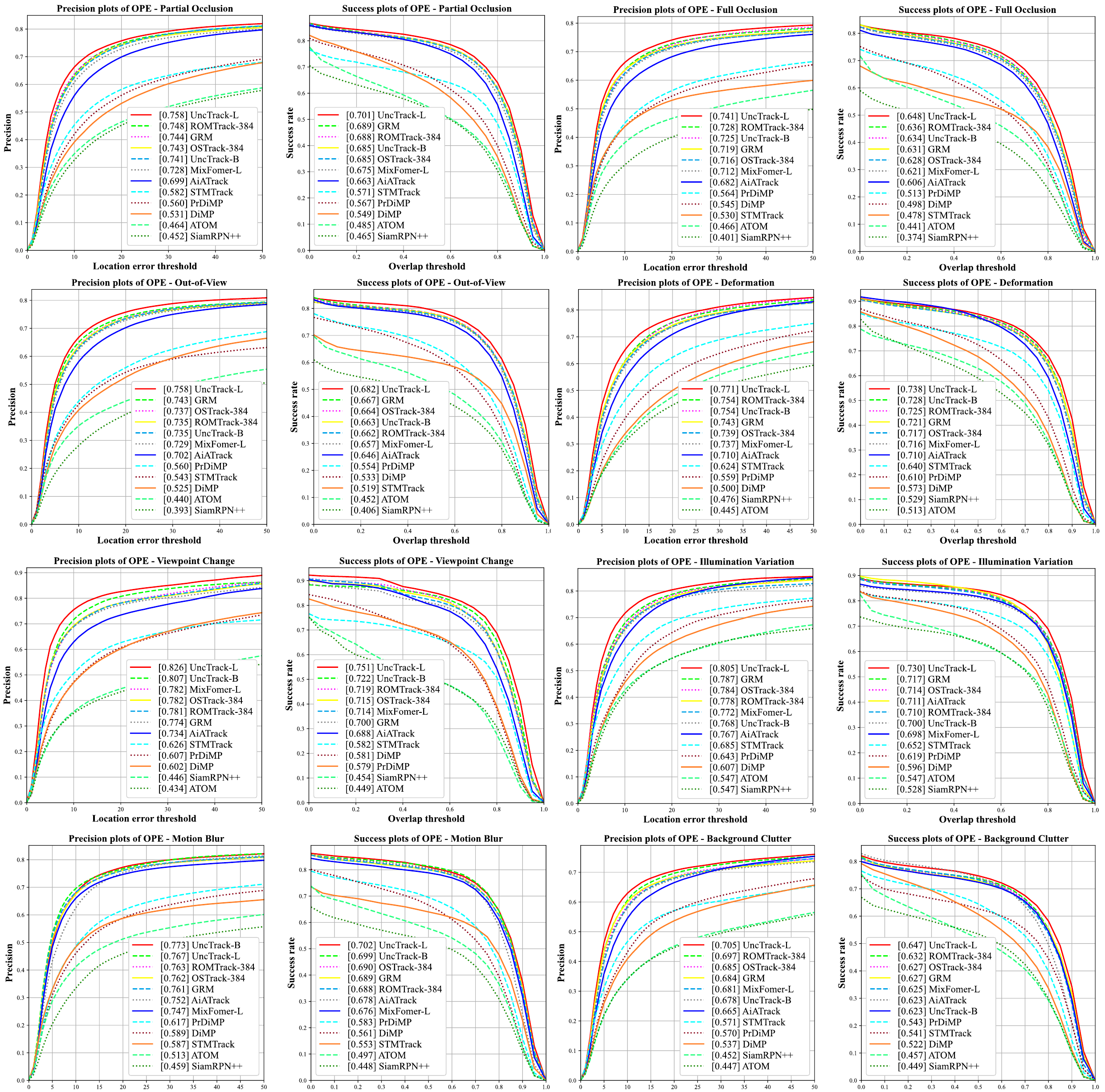}
\caption{Comparisons on LaSOT dataset with different challenging aspects: partial occlusion, full occlusion, out-of-view, deformation, viewpoint change, illumination variation, motion blur and background clutter. }
\label{fig:DIFF_ATTR}
\end{figure*}

\begin{figure*}
\centering
\includegraphics[width=6.2in]{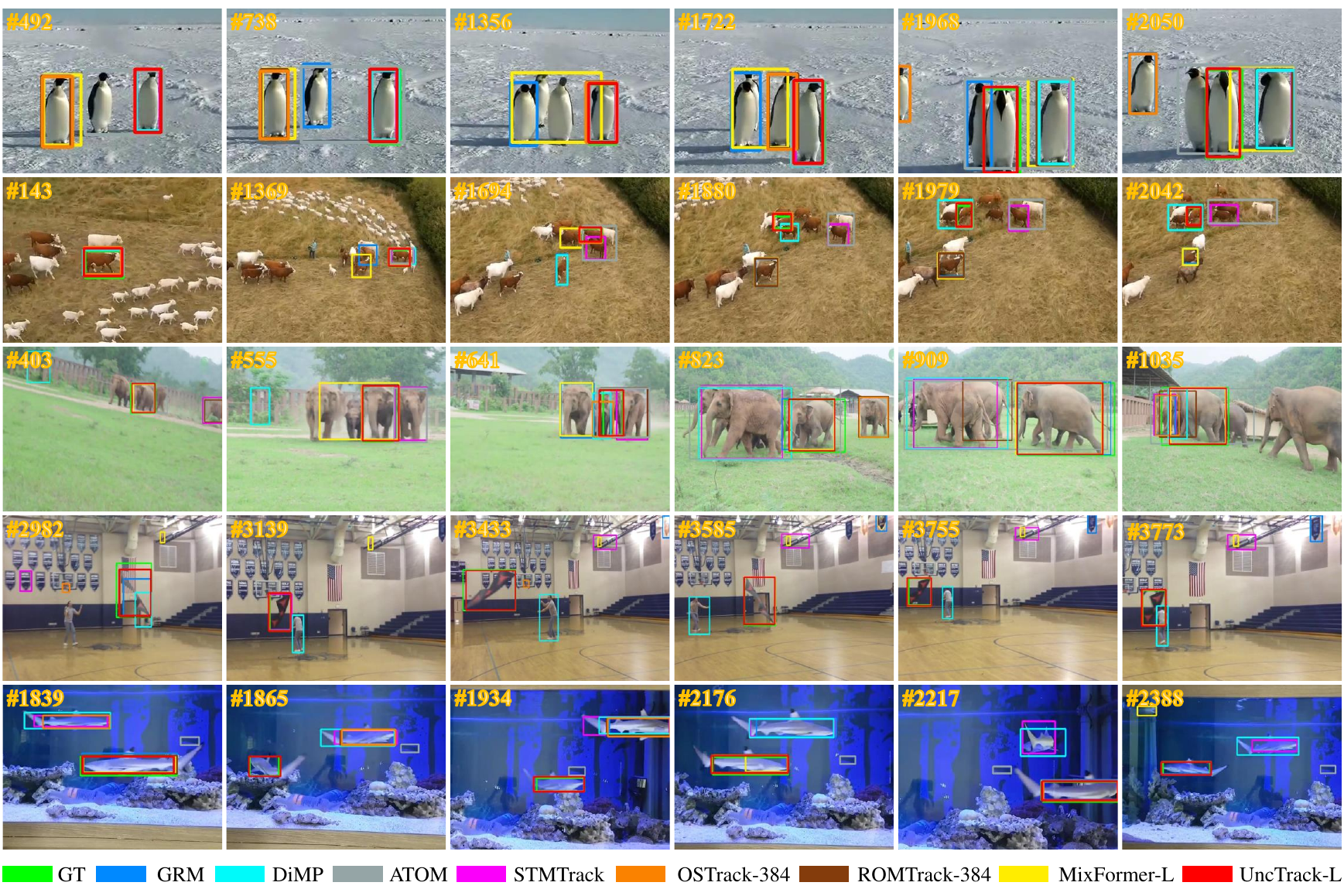}
\caption{Qualitative comparison of UncTrack against other state-of-the-art trackers in five challenging sequences of LaSOT dataset, including bird-2, cattle-13, elephant-12, kite-6 and shark-2.}
\label{fig:show_results}
\end{figure*}

\begin{figure*}[]
\centering
\includegraphics[width=6.2in]{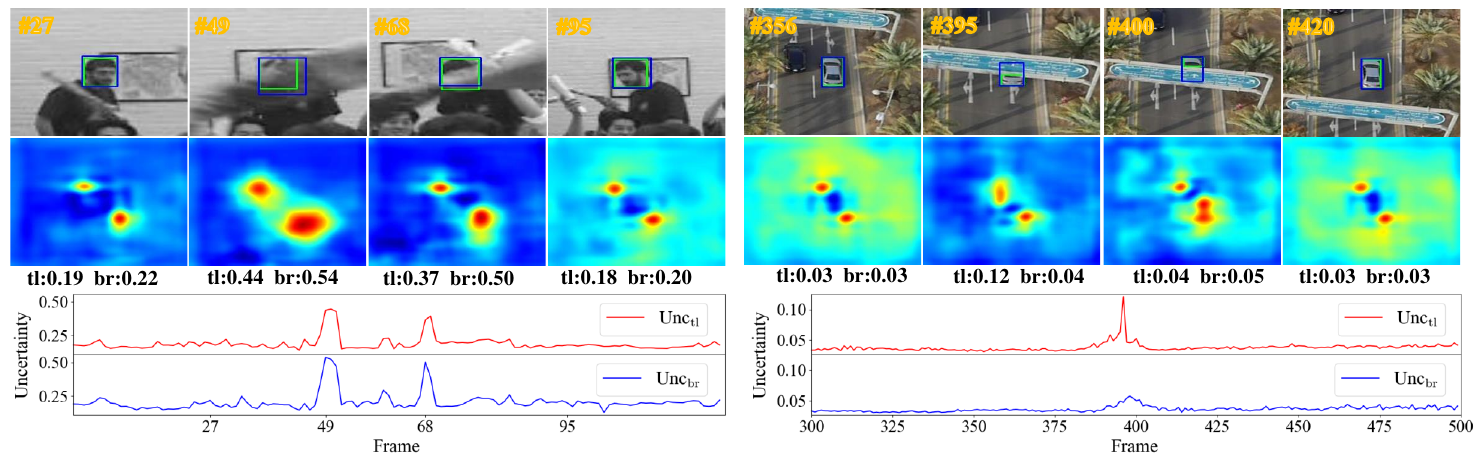}
\caption{Visualization of the localization heatmaps and the predicted uncertainties of the corner points in challenging sequences, the video frame with the highest uncertainty score indicates the unreliable output at this timestep.}
\label{fig:uncertain_vis}
\end{figure*}

\subsection{Related Attributes Evaluation}

We present deeper analyses of the proposed method in several challenging scenarios, including partial occlusion, full occlusion, out-of-view, and deformation, etc. This experiment is conducted on LaSOT dataset using precision and success rate plots for fair comparison, the quantitative plots and visualized results are shown in Fig. \ref{fig:DIFF_ATTR} and Fig. \ref{fig:show_results}, respectively.

\subsubsection{Partial Occlusion} Partial occlusion is a challenging characteristic of VOT due to the incomplete appearance modeling. In this scenario, UncTrack-L obtains the precision and success rate scores of $(0.758, 0.701)$, which outperforms other transformer trackers like ROMTrack-384 and MixFormer-L, etc. For example, in the challenging sequence of \emph{bird-2} with significant partial occlusion, UncTrack-L can estimate the target's location precisely while MixFormer fails in this case.

\subsubsection{Full Occlusion} As for the videos with more challenging full occlusion, the proposed UncTrack is also available for maintaining reliable trajectories. In these videos, the UncTrack-L obtains the best precision and success rate scores of $(0.741, 0.648)$, the UncTrack-B also achieves the competitive performance with the corresponding scores of $(0.725, 0.634)$. Compared with other methods such as GRM and OSTrack-384, our method achieves significant performance gain in this scenario. 


\subsubsection{Out-of-View} Videos with out-of-view attributes are also very challenging as trackers can hardly maintain accurate prediction if the target's moving trajectory becomes inconsistent. We can see that UncTrack-L achieves the highest scores in both precision and success plots. It obtains the precision score of 0.758 and success rate score of 0.682. UncTrack-B obtains the relative scores of 0.735 and 0.663, respectively. Such results verify that our UncTrack can effectively track the target objects even in the case that the objects disappear in a clip of video frames.

\subsubsection{Object Deformation} We observe our method is also effective in videos where objects undergo deformation. UncTrack-L ranks at the top place with the precision score of 0.771 and success rate score of 0.738. UncTrack-B achieves the performance of 0.754 and 0.728 in terms of precision and success rate metrics. These results indicate that the localization uncertainty mechanism in UncTrack enables it to adaptively discard unreliable samples, thus maintaining high-quality localization. Consequently, the proposed UncTrack demonstrates greater robustness against object deformation.

\subsubsection{Motion Blur} Motion blur introduces significant localization uncertainty during online tracking. Due to the effectiveness of PMN, the proposed UncTrack-L and UncTrack-B obtain the highest precision/AUC performance in the corresponding videos with motion blur attribute. We notice that even the lightweight version UncTrack-B can surpass AiATrack or MixFormer-L by $2.1\%$ and $2.3\%$ in terms of AUC.

\subsubsection{Background Clutter} For the videos with background clutter. UncTrack-L achieves the precision score of 0.705 and success rate score of 0.647. Compared with the SOTA methods like OSTrack-384 and MixFormer-L, UncTrack-L obtains the precision gains of $2\%$ and $2.4\%$, respectively.

\subsection{Heatmap and Uncertainty Visualization}
We explain how our UncTrack can effectively measure the uncertainty in challenging scenarios. Fig. \ref{fig:uncertain_vis} visualizes the localization heatmaps and the predicted uncertainties of the corner points at different timesteps. In the first video sequence, we find that UncTrack presents accurate localization response at the $27$th frame, and the uncertainty scores of the top-left and bottom-right corners are not high. When the full occlusion occurs in the $49$th and $68$th frames, the localization heatmaps become noticeably noisy, and the uncertainty outputs of these two corners increase significantly. The similar phenomenon can also be found in the second sequence, where the partial occlusion of the tracking object occurs in the $395$th frame. We can observe the uncertainty output of the top-left corner increases rapidly. These visualization analyses reveal that our UncTrack can accurately capture the corners' localization uncertainty. By further constructing reliable target-specific representation using PMN, UncTrack can be more robust against the challenging background distractors.

\begin{figure}[]
\centering
\includegraphics[width=3.4in]{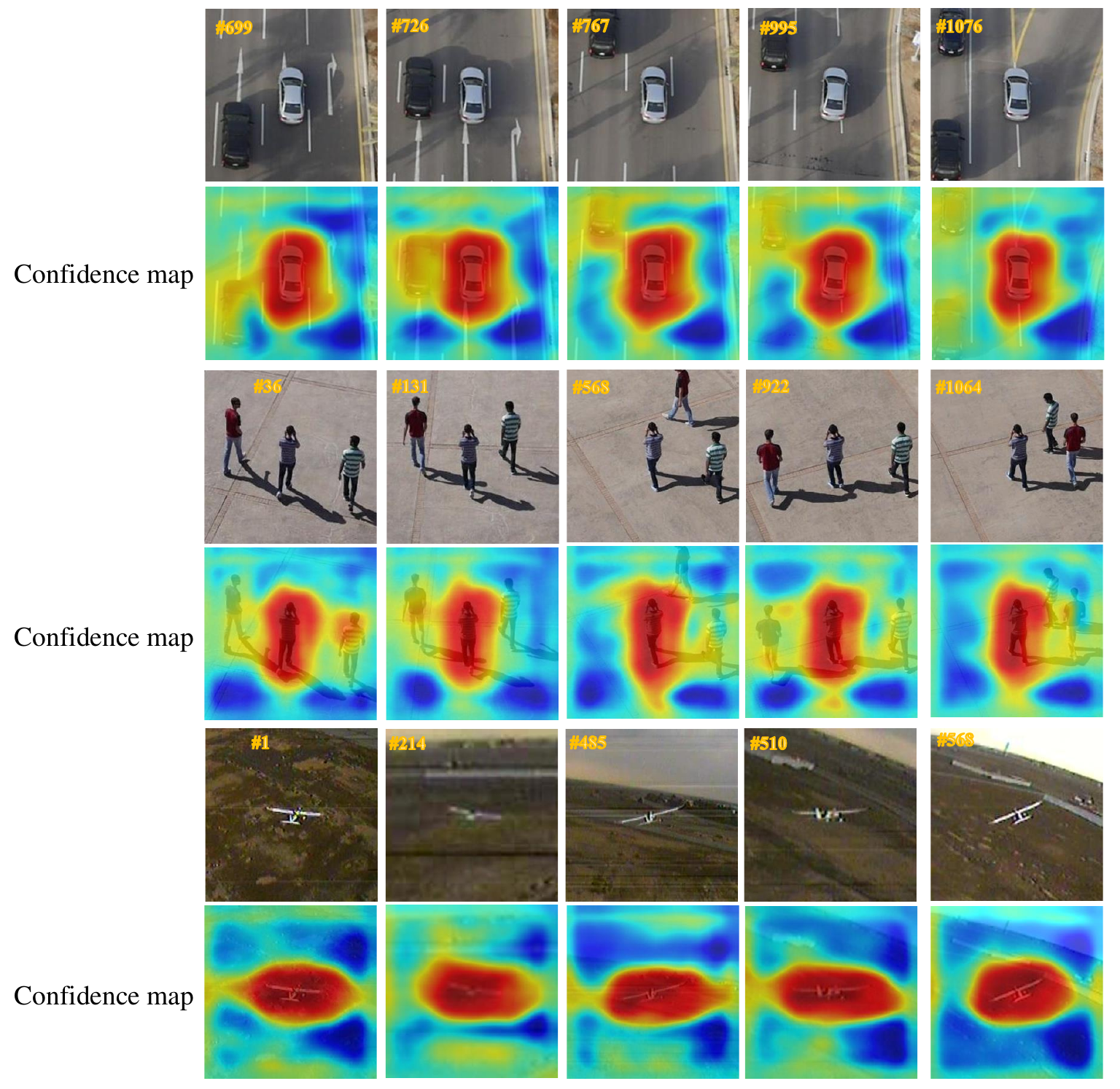}
\caption{Visualization of the transformed localization reliability in the proposed confidence inversion module (CIM).}
\label{fig:uncertainty_heatmap_r3}
\end{figure}

\subsection{Comparison with Other Online Trackers.}
The online updating mechanism has been utilized in several transformer-based online trackers, \emph{e.g.} MixFormer V2 \cite{DBLP:conf/nips/CuiSWW23} and STARK \cite{DBLP:conf/iccv/0002PF0L21}. Specifically, STARK adopts a three-layer perceptron to update the target template online. MixFormer V2 uses SPM to select reliable online templates. Both of these methods focus on cropping the local patches online to construct high-quality templates by positive/negative sample classification, while the temporal motion consistency is entirely ignored. Nevertheless, our UncTrack encodes the localization uncertainty in the continuous video frames as prototype representation. The position-aware prototypes are updated within the prototype memory bank using the temporal read-write operation, leading to more robust motion prediction in challenging scenarios. As shown in Fig. \ref{fig:uncertainty_heatmap_r3}, the output uncertainty heatmaps in ULD of UncTrack are sent into the Confidence Inversion Module (CIM), which serves as an indicator to measure whether each pixel belongs to the target object or not.

\begin{figure}[]
\centering
\includegraphics[width=3.4in]{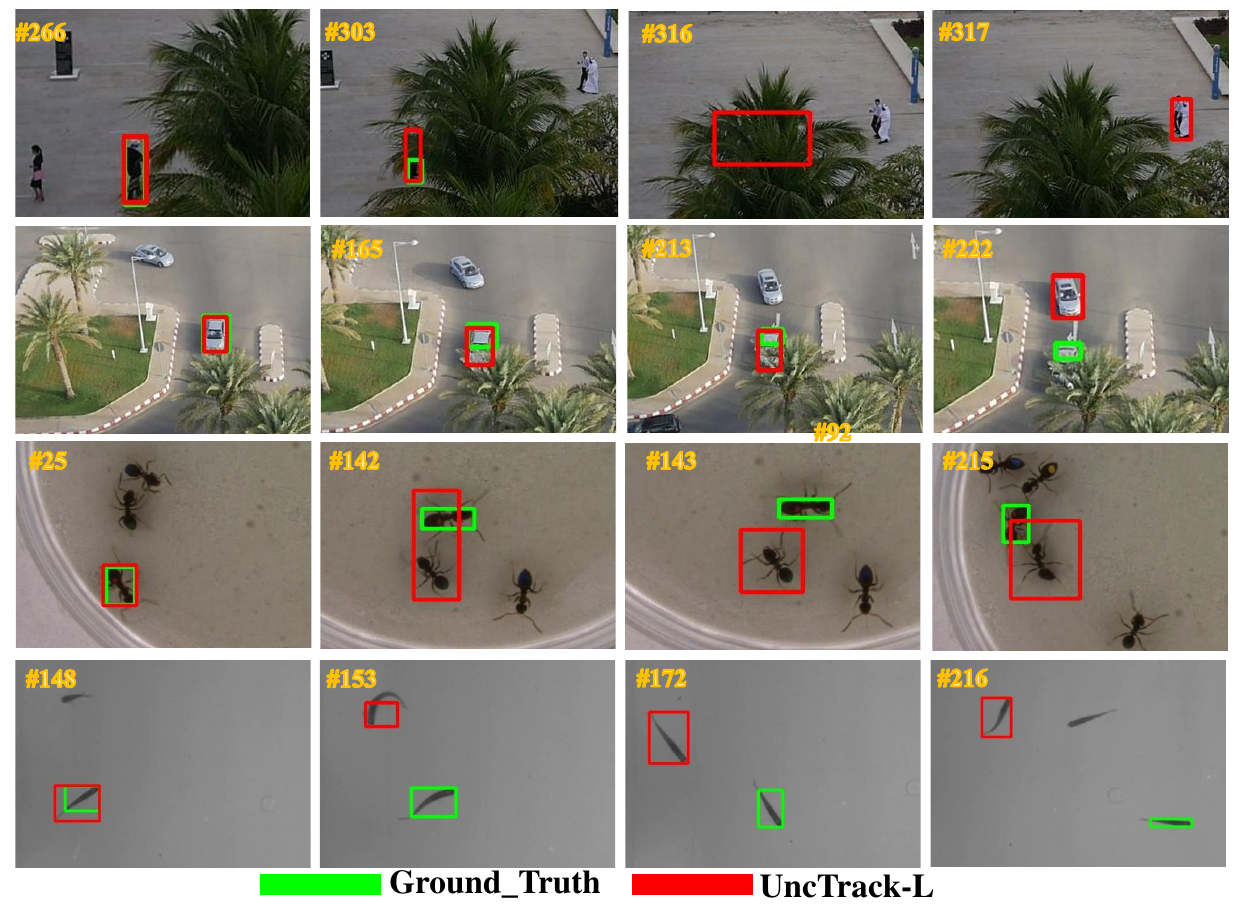}
\caption{The failure cases of UncTrack-L on UAV123 and VOT2020 datasets. }
\label{fig:failure_case}
\end{figure}

\noindent\textbf{Failure Cases.}
The proposed UncTrack mainly focuses on learning the corner-based localization uncertainty, making it well-suited to handle the challenging attributes like partial occlusion, out-of-view and deformation, etc. However, a crucial limitation is that the target-aware discriminative capability of UncTrack is still significantly determined by the ViT backbone. Therefore, if two objects are located closely and share the similar visual appearance, while the tracking target is occluded by background distractors (e.g., trees or advertising boards), UncTrack may fail to distinguish the similar object. The failure cases are shown in Fig. \ref{fig:failure_case}.

\section{Conclusions}
In this paper, we propose a novel uncertainty-aware transformer tracker termed UncTrack, which exploits the localization uncertainty information for accurate target state inference. UncTrack flattens the paired template-search images into token patches and performs feature interaction using a transformer encoder to establish discriminative feature representation. Then the output features are passed into an uncertainty-aware localization decoder (ULD) to predict the target state and the localization uncertainty. The uncertainty output and the template-search features are jointly sent into a prototype memory network (PMN) to identify the reliability of target state prediction. The samples with high confidence are fed back into the prototype memory bank for memory updating. Extensive experiments demonstrate that our method achieves superior performance on the public benchmarks.


\ifCLASSOPTIONcaptionsoff
  \newpage
\fi



%



\bibliographystyle{IEEEtran}

%




\end{document}